\documentclass[10pt,twocolumn,letterpaper]{article}

\usepackage[final]{sty/cvpr}

\usepackage[utf8]{inputenc} 
\usepackage[T1]{fontenc}    
\usepackage{url}            
\usepackage{booktabs}       
\usepackage{amsfonts}       
\usepackage{nicefrac}       
\usepackage{microtype}      
\usepackage[dvipsnames]{xcolor}         
\usepackage{multirow}
\usepackage{dcolumn}
\usepackage{xspace}
\usepackage{makecell}
\usepackage{amsmath}
\usepackage{graphicx}
\usepackage{amssymb}
\usepackage{bbm}
\usepackage{paralist}
\usepackage{tabularx}
\usepackage{colortbl}
\usepackage[ruled,vlined]{algorithm2e}
\definecolor{citecolor}{HTML}{2980b9}
\definecolor{linkcolor}{HTML}{c0392b}
\usepackage[pagebackref=true,breaklinks=true,colorlinks,bookmarks=false,citecolor=citecolor,linkcolor=linkcolor]{hyperref}
\usepackage{amsmath,bm}
\usepackage{lscape} 

\usepackage{pifont}

\usepackage[pagebackref=true,breaklinks=true,colorlinks,bookmarks=false]{hyperref}

\renewcommand{\paragraph}[1]{\vspace{0pt}\noindent\textbf{#1}}

\newcommand{\thor}{\mbox{\sc{AI2-Thor}}\xspace}
\newcommand{\taskshort}{\mbox{\sc{ShIf}}\xspace}
\newcommand{\tasklong}{\mbox{\sc{LhIf}}\xspace}
\newcommand{\taskiqa}{\mbox{\sc{IQA}}\xspace}
\newcommand{\taskexp}{\mbox{\sc{ExIn}}\xspace}

\newcommand{\net}{\mbox{Hierarchical Interactive Network}\xspace}
\newcommand{\netsmall}{\mbox{\sc{Hint}}\xspace}
\newcommand{\netbase}{\mbox{\sc{Flat}}\xspace}
\newcommand{\asc}{\mbox{\sc{Asc}}\xspace}
\newcommand{\tabrow}[1]{\small\texttt{\textcolor{blue}{#1}}}
\newcommand{\tabrowb}[1]{\scriptsize\texttt{\textcolor{blue}{#1}}}

\definecolor{rowgray}{gray}{0.85}
\definecolor{mygray}{gray}{0.4}

\date{}

\def\cvprPaperID{9298} 
\def\confName{CVPR}
\def\confYear{2022}

\begin{document}

\title{ASC me to Do Anything: \\ Multi-task Training for Embodied AI}

\author{Jiasen Lu \quad Jordi Salvador \quad Roozbeh Mottaghi \quad Aniruddha Kembhavi\\
\\
PRIOR @ Allen Institute for AI\\
{\tt\small \{jiasenl, jordis, roozbehm, anik\}@allenai.org}
}

\maketitle
\begin{abstract}
Embodied AI has seen steady progress across a diverse set of independent tasks. While these varied tasks have different end goals, the basic skills required to complete them successfully overlap significantly. In this paper, our goal is to leverage these shared skills to learn to perform multiple tasks jointly. We propose Atomic Skill Completion (\asc), an approach for multi-task training for Embodied AI, where a set of atomic skills shared across multiple tasks are composed together to perform the tasks. The key to the success of this approach is a pre-training scheme that decouples learning of the skills from the high-level tasks making joint training effective. We use \asc~to train agents within the \thor~environment to perform four interactive tasks jointly, and find it to be remarkably effective. In a multi-task setting, \asc~improves success rates by a factor of 2x on Seen scenes and 4x on Unseen scenes compared to no pre-training. Importantly, \asc~enables us to train a multi-task agent that has a 52\% higher Success Rate than training 4 independent single task agents. Finally, our hierarchical agents are more interpretable than traditional black box architectures.

\end{abstract}

\section{Introduction}

Embodied AI (E-AI) researchers have long sought to develop agents that can perform complex tasks within visual environments -- tasks that require navigating around an environment~\cite{Anderson2018OnEO,Batra2020ObjectNavRO}, interacting~\cite{Shen2020iGibsonAS,Zeng2021PushingIO,Zhu2017VisualSP,Batra2020RearrangementAC,Weihs2021VisualRR} and manipulating~\cite{Ehsani2021ManipulaTHORA,Xiang2020SAPIENAS} with objects that lie within it, following instructions~\cite{anderson2018vision,ALFRED20} and engaging with other agents~\cite{Jain2019TwoBo,Jain2020ACS} or humans via QA~\cite{gordon2018iqa,embodiedqa}. While steady progress has been made towards this ambitious goal, particularly in simulated worlds~\cite{Shen2020iGibsonAS,savva2019habitat,Kolve2019AI2THORAn,Xiang2020SAPIENAS,Gan2020ThreeDWorldAP}, most work today focuses on training agents to perform a single task.

Evidence across this large body of research suggests that: (a) Present day learning algorithms are very inefficient -- perfecting simpler tasks such as point goal navigation can require more than a billion frames of experience~\cite{Wijmans2020DDPPOLN}, (b) The best performing methods are not as effective for long horizon tasks as well as tasks that involve rich interactions with the world and larger action spaces~\cite{Weihs2021VisualRR}, and (c) As tasks get more complex, the generalization capability of these methods to unseen environments is quite poor~\cite{ALFRED20}. How do we move towards developing effective multi-task E-AI agents, when training single task ones continues to be challenging?

In this paper, we propose an approach to jointly train multiple embodied tasks specified by natural language. Training these tasks directly is quite challenging. However, the compositional nature of natural language allows us to decompose the tasks into smaller easy-to-train parts that are shared across multiple tasks. For example, given two different tasks ``\textit{put a plunger in cabinet}.'' and ``\textit{what is the color of plunger?}'', the agent must execute ``\textit{find plunger}'' first, which is essentially an object navigation task. Our central idea is to pre-train such smaller tasks, referred to as \emph{atomic skills} and later compose them to perform the more complex downstream task. More specifically, we pre-train a low-level skill executing policy on a set of atomic skills applicable to diverse downstream tasks, and then, train a high-level skill invocator on downstream tasks.

Our method, referred to as Atomic Skill Completion (\asc), has several advantages: (1) Since \asc pre-trains on atomic skills instead of long-horizon downstream tasks, it pre-trains fairly quickly. (2) These skills are short horizon and relatively simple to learn and thus generalize better to unseen environments. (3) The atomic skills are chosen to be generic and can be chained to perform many downstream E-AI tasks. (4) Since high-level task invocation is now decoupled from low-level skill execution, the learning of high-level policies is hugely simplified, leading to the creation of effective multi-task embodied agents.

\begin{figure*}[tp]
    \centering
    \includegraphics[width=0.95\textwidth]{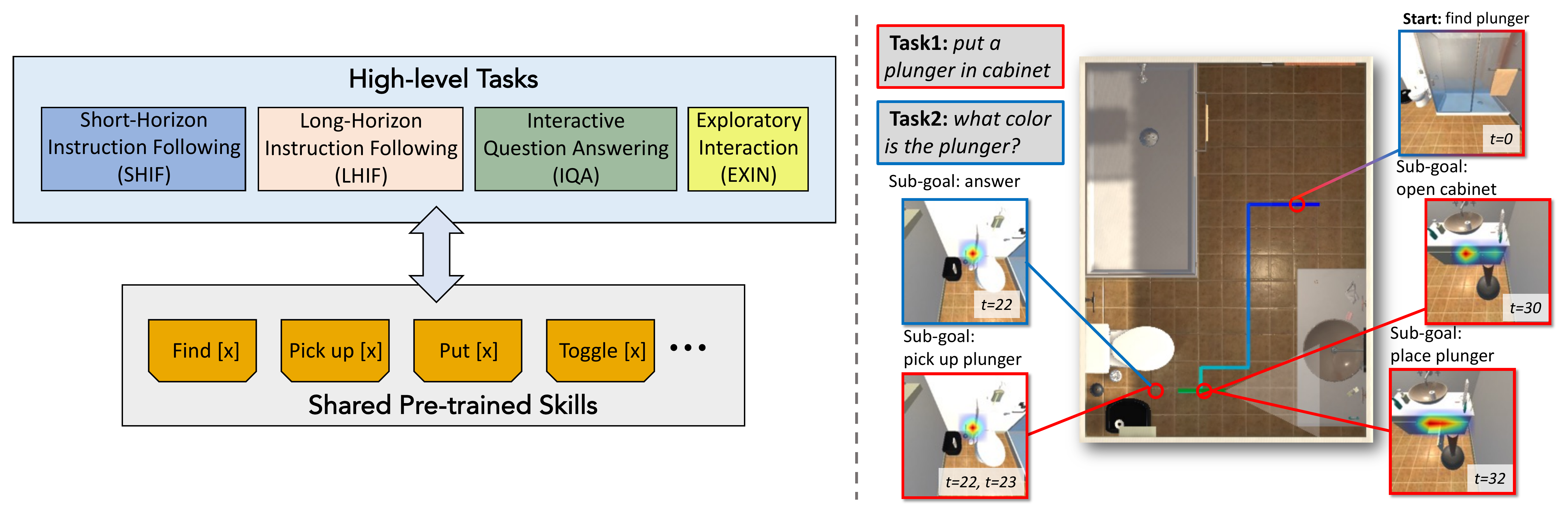}
    \caption{We consider four high-level tasks that require navigation and object interaction. Skills are meaningful short sequences of primitive actions that are pre-trained and shared among tasks. On the right, we show an example of an episode by our multi-task agent for the long-horizon instruction following task of \emph{put a plunger in cabinet.} and an interactive question answering task of \emph{what color is plunger?}}
    \label{fig:teaser}
\end{figure*}

We pre-train our agent within the \thor environment with 110 object classes, 13 actions and continuous parameterization of the interaction skills. We consider 9 atomic skills common across a wide range of higher level tasks. They range from navigation skills such as \texttt{find X} to interactive skills such as \texttt{slice X} and the answering skill \texttt{answer}. We then jointly train the agent to perform 4 challenging tasks (Figure~\ref{fig:teaser}) -- (1) Short Horizon Instruction Following (\taskshort), (2) Long Horizon Instruction Following (\tasklong), (3) Interactive Question Answering (\taskiqa) and (4) Exploratory Interaction (\taskexp) and measure performance on Seen and Unseen environments. Given the interaction-heavy nature of these tasks, we consider two interaction modes -- \emph{Standard} whereby the agent must predict a bounding box that overlaps with the target object, and \emph{Hard} which requires the agent to accurately predict a point within the object that it wishes to interact it. 

Our results show that pre-training the agent via \asc leads to large improvements across all four tasks.
In the Standard setting for multi-task training, \asc improves Success Rates (averaged across all tasks) from 15.1 $\rightarrow$ 41.9 for Seen and 4.3 $\rightarrow$ 16.2 for Unseen; in the Hard setting, the improvements are as dramatic -- from 20.3 $\rightarrow$ 39.8 for Seen and 4.0 $\rightarrow$ 19.3 for Unseen. 
In the absence of pre-training, multi-task training results in a drop as compared to single task training for Unseen scenes (5.0 $\rightarrow$ 4.0), but when using \asc, multi-task training provides a large boost (10.9 $\rightarrow$ 19.3).
We also find that our multi-task agent with \asc performs comparably in the Hard setting vs the Standard setting -- 41.9 vs 39.8 Seen and 16.2 vs 19.3 Unseen, reflecting the ability of our agent to precisely predict the locations of target objects that it needs to interact with in the scenes. Finally, we find our model more interpretable than traditional Embodied AI solutions, since we can observe the sub-goals, target object and pixel locations for interaction output at each time step, enabling us to monitor progress and explain some successes and errors.

\section{Related Work}

\noindent \textbf{Multi-task learning.} An ultimate goal of AI research is to build systems that can perform multiple tasks simultaneously. There are several previous works in computer vision \cite{Kokkinos2017UberNetTA,Eigen_2015_ICCV,Ren_2018_CVPR,Liu2019EndToEndML,Mallya2018PackNetAM,Sener2018MultiTaskLA,Misra2016CrossStitchNF}, natural language understanding \cite{collobert2008,liu-etal-2015-representation,liu2019mt-dnn,mccann2018natural,clark-etal-2019-bam,khashabi2020unifiedqa} and vision \& language \cite{Lu_2020_CVPR,gpv,unit,omninet,learn_them_all} domains that aim to handle multiple tasks simultaneously and address the issues that arise when tackling different tasks together. However, the visual embodied research works primarily focus only on a single specific task \cite{Zhu2017VisualSP,Wijmans2020DDPPOLN,Wortsman2019LearningTL,embodiedqa}. There are a few previous works that consider multi-task scenarios in the E-AI domain. For example, \cite{ijcai2020-338} transfer the knowledge of words and their grounding across two navigation tasks and \cite{DBLP:journals/corr/abs-2003-00443} share parameters for language encoding and policy between two vision and language navigation tasks. Prior work in training multi-task agents is either operate in a grid-world environment \cite{andreas2017modular}, or environments with limited complexities, such as mine-craft to stack blocks \cite{shu2017hierarchical} or ViZDoom with a single room including 5 objects \cite{ijcai2020-338}. We focus on long-horizon tasks that involve object interaction and state changes in addition to navigation. Furthermore, we show the effectiveness of pre-training of skills for learning  different tasks jointly. 

\noindent \textbf{Pre-training.} Pre-training strategies using supervised or unsupervised methods have proven to be effective in terms of learning efficiency and performance for downstream tasks in computer vision \cite{girshick2014rich,mahajan2018exploring,sun2017revisiting,ghadiyaram2019large} and NLP \cite{elmo,radford2019language,Devlin2019BERTPO,gpt3}. Recently, pre-training methods have become popular in the E-AI domain. \cite{du2021curious} jointly learn a policy and visual representations and show transfer to downstream navigation tasks. \cite{Wijmans2020DDPPOLN} propose a pre-training scenario that provides massive performance gains for the task of point navigation. \cite{midLevelReps2018,pmlr-v100-sax20a} use mid-level tasks such as depth and room layout estimation for learning representations that enable fast and more generalizable learning of downstream tasks. \cite{ramakrishnan2021environment,oord2019representation} have used contrastive predictive coding ideas to pre-train networks for downstream navigation tasks. \cite{Gordon_2019_ICCV} explore pre-training using auxiliary tasks. \cite{Li2020UnsupervisedRL} learn skills for navigation via meta-reinforcement learning. Most of these works consider navigation as the downstream task. In contrast we consider tasks that involve object interaction. Moreover, we propose a pre-training strategy for a set of atomic skills that are composed in a hierarchical fashion.  

\noindent \textbf{Hierarchical planning.} There is a rich history of hierarchical planning for performing different types of tasks \cite{Mcgovern01automaticdiscovery,goel2003subgoal,NIPS2016_f442d33f,pmlr-v54-fruit17a, le2018hierarchical}. Here, we mention a few approaches that are most relevant to ours. \cite{das2018neural,gordon2018iqa} propose a hierarchical architecture for embodied question answering. \cite{xiali2020relmogen} address the problem of sub-goal generation and finding a sequence of actions to reach the sub-goal. \cite{Jiang2019LanguageAA} use language as an abstraction to break down a complex task. \cite{Nair2020Hierarchical} generate image sub-goals conditioned on an image goal and use the sub-goals for planning. \cite{DBLP:journals/corr/FlorensaDA17} learn skills using intrinsic motivation to speed up learning downstream tasks that share a common structure. \cite{minec} address the problem of reusing and transferring knowledge from one task to another. \cite{pmlr-v70-oh17a} propose an analogy-making objective to generalize to unseen tasks and also a method for estimating the time-scale of sub-tasks. \cite{eysenbach2018diversity} address the problem of learning skills without a reward. These approaches have one or more of the following limitations: they do not train for multiple \emph{distinct} tasks, they focus on simple tasks that do not require simultaneous navigation and object interaction, they use the same environment for train and test, or they do not consider high-dimensional visual input. 









\section{Problem Statement}

An important goal of Embodied AI research is to develop agents that can perform a wide variety of diverse tasks. However, training multiple tasks at once is quite challenging since each task has a different success criteria, tasks often have different output structures, and training time becomes insurmountable. We propose pre-training agents on a shared set of skills that are core components of the target downstream tasks. As we show in our experiments, this improves training for complex interactive tasks and enables us to jointly train for multiple tasks.

\textbf{Skills.} The skills used to pre-train our agent are defined manually and correspond to semantically meaningful interactions with the environment that require very short sequences of primitive actions. We consider nine skills that involve navigation, interaction and answering questions. For example, \textit{go to} and \textit{turn on} are two example skills that we consider. These skills are part of a wide variety of interactive tasks.

\textbf{Tasks.}
We consider four target tasks: (1) \emph{short-horizon} instruction following; these tasks typically require a few skills (e.g., \emph{clean tomato}, which requires putting the tomato in a sink and turning on the faucet). (2) \emph{long-horizon} instruction following; these tasks span a longer horizon compared to short-horizon tasks and are inspired by the seven tasks defined in ALFRED~\cite{ALFRED20}, where only the high-level goal is available to the agent (as opposed to step-by-step instructions). An example is \emph{put the fork in the cup and move them to the kitchen counter}. (3) \emph{interactive question answering}, which is inspired by previous works of~\cite{gordon2018iqa,embodiedqa}. The goal is to answer questions that require interaction with a scene. In this paper, the questions query the visual state or quantity of objects (e.g., \emph{Is the fridge open?},  \emph{How many eggs are in the fridge?}). (4) \emph{exploratory interaction}; this task requires a long exploration phase until it finds the target object with which it needs to interact. This task shares similarities with the first two tasks, but it is more navigation-heavy compared to those. An example task is \emph{pickup the apple}, which requires invoking the \emph{navigation} skill multiple times (in case of failures) to reach the apple and then invoking the interaction skill \emph{pick up} to pick up the apple). 

\begin{figure*}[ht!]
    \centering
    \includegraphics[width=0.95\textwidth]{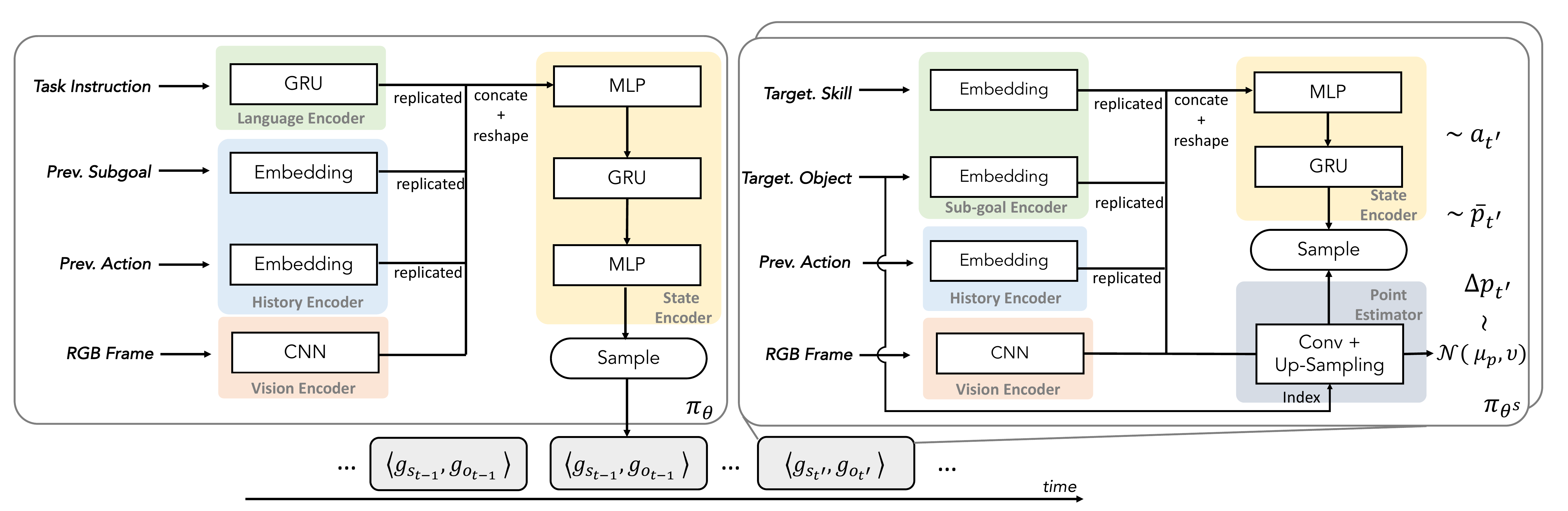}
    \caption{\textbf{\net (\netsmall).} The high-level policy $\pi_\theta$ generates a sub-goal $g_t$ at each time $t$ (shown in grey at the bottom of the figure). The right panel shows a sub-policy $\pi_{\theta^s}$ for an interaction skill. Each sub-policy outputs a primitive action and a point on the image for interacting with objects.} 
    \label{fig:model}
\end{figure*}

\textbf{Environment.}
We use AI2-THOR~\cite{Kolve2019AI2THORAn}, a visually rich interactive framework, for performing our tasks. Following~\cite{ALFRED20}, no prior knowledge about the environment (e.g., a map) or additional sensors (e.g., depth cameras and GPS sensors) are available to the agent. We consider 110 object classes (37 of which are receptacle object classes) across 112 different indoor scenes. The environments provide multiple variations of each object class with different shapes, textures and colors. 

\textbf{Agent.} 
To complete a task instance $T$, at each time step $t$, the agent observes an egocentric RGB image $v_t$ as input and takes action $a_t$, which can be a navigation action (e.g., \emph{move ahead}, \emph{rotate right}), an object interaction action (e.g., \emph{pick up}, \emph{slice}) or an answer action (e.g., \emph{yes}, \emph{3}). The full list of actions is provided in the appendix. At each time step the agent also produces a coordinate $p=(x,y)$ on the image plane to indicate the object that will be interacted with. For instance, if the agent wishes to pick up a bowl, it issues the action \emph{pick up} along with the coordinates of a pixel within the segment corresponding to the bowl. The agent's objective is to learn a policy that can successfully complete task $T$. We consider a hierarchical policy $\pi_\theta$, which decomposes a task $T$ into multiple skills and dynamically selects sub-policies $\pi_{\theta^s}$ (corresponding to the desired skills) to execute.

\textbf{Continuous Interaction Parameterization.} A common practice for specifying target objects for interaction is to predict the target segmentation mask~\cite{ALFRED20} or bounding box~\cite{gordon2018iqa}, then compare this to the ground truth segmentation masks provided by the simulator, select the \emph{interactable} object with the most overlap and then use that as the target object. We name this setting \emph{Standard} and also explore a more challenging continuous interaction parameterization, named \emph{Hard}, a more realistic setting that no knowledge of the groundtruth is available to the agent. We interact with the environment by predicting a point $p=(x, y)$ on the RGB image. If the point is on an object and the object is within the range of interaction, the agent can interact with the object. Otherwise, the interaction action will fail. 

\section{Multi-task Training}

Given a task instance $T$ specified by language, our agent predicts a sequence of skills and executes them to achieve the desired goal. Our aim is to train for multiple different tasks jointly. The tasks might have conflicting goals. For example, some tasks heavily rely on navigation within a scene, while others require long sequences of object interaction actions. Furthermore, the span of the tasks can vary significantly. Some tasks can be performed by executing a short sequence of actions, while others require a longer sequence. This imbalance makes joint training unstable. To tackle these challenges, we propose a hierarchical policy, which relies on a pre-training strategy for a set of skills. In this paper, we focus on instruction following, question answering, and exploration tasks, but our proposed framework is applicable to a larger set of tasks that can be specified using language and that can be accomplished using a shared set of skills.

We first present our hierarchical policy with continuous interaction parameterization, as shown in Fig.~\ref{fig:model}. Then, we describe the pre-training strategy for the skills. Finally, we describe how the various modules are combined and trained with a recovery planner. 

\subsection{Hierarchical Policy}
\label{sec:hier}

Our hierarchical policy, which we name Hierarchical Interactive Network (\netsmall), decomposes the task instance $T$ into multiple sub-goals. Let $v_t$ and $a_t$ denote the observation and a primitive action (e.g., \emph{turn right}) at time $t$, respectively, and $g_t=\langle g_{s_t},g_{o_t} \rangle$ denote a sub-goal, where $s_t$ is the skill required to achieve the sub-goal $g$ and $o_t$ is the object required (if any) for that skill. The learning problem can be formulated as joint learning of a high-level policy $\pi_\theta: (T,a_{t-1},g_{t-1},I_t) \rightarrow g_t$ parameterized by $\theta$ and sub-policies $\pi_{\theta^s}: (g_t, a_{t-1},v_t) \rightarrow (a_t,p_t)$ parameterized by $\theta^s$ , where $p_t$ is an interaction point on the image or \texttt{None} if the sub-policy does not require to interact with an object. For example, $T$ can be ``Heat potato''. One of the sub-goals, $g_t$, will be ``Open Microwave'', where $g_{s_t}$ is the ``Open'' skill and $g_{o_t}$ is ``Microwave''. $p_t$ should be a point on the microwave so the agent can interact with it. 

\textbf{High-level Policy.}
The high-level policy is implemented as a single layer Gated Recurrent Unit (GRU). 
Given the task instance $T$, we first use a single layer GRU to extract the task embedding $\bm{z}^T$. We pass the current visual observation $v_t$ into a pre-trained ResNet18, producing an encoded convolutional image feature $\bm{z}_t^{\texttt{img}} \in \mathbb{R}^{d \times w \times h}$, where $d$ is the feature dimension, $w$ and $h$ is the size of covolutional feature map.

Besides the image feature and task feature, we also encode the last primitive action $a_{t-1}$ and last sub-goal $g_{t-1} = \langle g_{s_{t-1}}$, $g_{o_{t-1}} \rangle$ into embeddings $\bm{z}_{t-1}^{\texttt{act}}$, $\bm{z}_{t-1}^{\texttt{gs}}$ and $\bm{z}_{t-1}^{\texttt{go}}$ by linear projection. Following \cite{weihs2020allenact}, we use a multi-layer embedding network to encode $[\bm{z}^T,  \bm{z}_{t-1}^{\texttt{act}}, \bm{z}_{t-1}^{\texttt{gs}}, \bm{z}_{t-1}^{\texttt{go}}]$ into a compressed embedding $\bm{z}_t^c \in \mathbb{R}^d$. We further replicate $\bm{z}_t^c$ into $\mathbb{R}^{d\times w \times h}$ and concatenate with $\bm{z}_t^\texttt{img}$. This information is reshaped into a 1-d vector and used to update the hidden states $h_t$ of the GRU. The policy then produces a probability distribution over all the possible skills and target objects.
\begin{equation}
    g_{s_t}, g_{o_t} \sim \pi([\bm{z}_t^c, \bm{z}_t^{\texttt{img}}])
\end{equation}
In recent hierarchical models (e.g., \cite{das2018neural}), the high-level policy typically updates its hidden states and sub-goals only when the previous sub-goal is finished or after a fixed number of steps. In contrast, our high-level policy updates its hidden states and the sub-goal after each primitive action is taken. This enables the high-level policy to observe the whole trajectory. The left panel of Figure~\ref{fig:model} shows the high-level policy structure.
 
\textbf{Sub-Policies.} 
We have three distinct sub-polices in our experiments: navigation $\pi^{\texttt{nav}}$, interaction $\pi^{\texttt{act}}$ and question answering  $\pi^{\texttt{qa}}$. 
The navigation and interaction sub-policies are distributions over primitive actions $a_t$. We concatenate the last primitive action embedding $\bm{z}_{t-1}^{\texttt{act}}$ with sub-goal embedding sampled from the high-level policy $\bm{z}_{t}^{\texttt{gs}}$, $\bm{z}_{t}^{\texttt{go}}$ to generate $\bm{\hat{z}}_t^c$. 
\begin{equation}
    a_t \sim \pi^{\texttt{nav/act}}([\bm{\hat{z}}_t^c, \bm{\hat{z}}_t^{\texttt{img}}]),
\end{equation}
where $\bm{\hat{z}}_t^{\texttt{img}}$ is the covolutional image feature from a separate ResNet18 network. 

To interact with the target object $g_{o_{t}}$, the interaction sub-policy $\pi^{\texttt{act}}$ should point to a region of $g_{o_{t}}$ in the current observation. 
We treat the pointing process as an additional action to sample from the $\pi^{\texttt{act}}$. 
However, the action space for pointing is enormous -- the number of pixels in the image are often 10000+ and training with this action space is infeasible. 
Therefore, we use a combination of discrete and continuous action parameterization to effectively represent the pointing action. 

We first discretize the image into a $B \times B$ grid -- which is a smaller action space -- to obtain a rough estimate of the object location $\bar{p}$. Then, the discretization error can be recovered by sampling a continuous offset 
$\Delta p $ from a multivariate normal distribution with mean $\mu_p$ and variance $\nu$. We feed $\bm{\hat{z}}_t^{\texttt{img}}$ into three \{$3 \times 3$ Conv, BatchNorm, ReLU, $3 \times 3$ Conv, BatchNorm\} blocks, producing the augmented feature $\bm{\hat{z}}_t^{\texttt{aug}}\in \mathbb{R}^{d\times w \times h}$. We feed $\bm{\hat{z}}_t^{\texttt{aug}}$ into three $1 \times 1$ Convs to produce the discrete location of the target point and the mean and the variance of multivariate normal distribution.

The target point can be estimated as $ p = \bar{p} + \Delta p$:
\begin{equation}
    \bar{p} \sim \pi_{\texttt{p}}^{\texttt{act}}(\bm{\hat{z}}^{\texttt{aug}}), \; \; \; 
    \Delta p \sim \mathcal{N} (\mu_p, \nu)
\end{equation}
For the question answering sub-policy, we use a standard VQA model that encodes the question with a single-layer GRU and performs dot product-based attention between the question encoding and the convolutional image feature. The answer can be sampled from the distribution produced by $\pi^{\texttt{qa}}$. 

\newcommand{\band}{\rowcolor{gray!10}}
\begin{table*}[!ht]\footnotesize
\setlength\tabcolsep{2pt}
    \center
  \resizebox{\textwidth}{!}{
    \begin{tabular}{clccccccccccccccccc}
        \toprule
        & \multirow{3}{*}{Model} & \multirow{3}{*}{Training} & \multicolumn{2}{c}{\texttt{Goto}} & \multicolumn{2}{c}{\texttt{Pickup}}  & \multicolumn{2}{c}{\texttt{Put}}  & \multicolumn{2}{c}{\texttt{ToggleOn}}  & \multicolumn{2}{c}{\texttt{ToggleOff}}  & \multicolumn{2}{c}{\texttt{Open}}  & \multicolumn{2}{c}{\texttt{Close}} & \multicolumn{2}{c}{\texttt{Slice}} \\
        \cmidrule(r){4-5}
        \cmidrule(r){6-7}
        \cmidrule(r){8-9}
        \cmidrule(r){10-11}
        \cmidrule(r){12-13}
        \cmidrule(r){14-15}
        \cmidrule(r){16-17}
        \cmidrule(r){18-19}

        & & & seen & unseen & seen & unseen & seen & unseen & seen & unseen & seen & unseen & seen & unseen& seen & unseen & seen & unseen\\
        \midrule
        \tabrow{1} & Interact & TF  & - & - & 8.3 & 5.0 & 25.0 & 20.0 & 54.2 & 41.9 & 53.5 & 36.2 & 44.0 & 19.4 & 55.6 &  31.2 & 22.2 & 12.5  \\
        \tabrow{2} & Interact & SF & - & - & 16.7 & 10.0 & 29.9 & 19.4 & 66.7 & 49.6 & 53.6 & 49.6 & 48.6 & 42.5 & 64.6 & 49.4 & 27.8 & 15.0 \\
        \tabrow{3} & Interact & MIX & - & - & 22.2 & 10.6 & \textbf{42.4} & 23.7 & 69.4 & 51.9 & 66.6 & 43.8 & \textbf{61.1} & 31.2 &  \textbf{71.5} & 45.0 & 33.3 & \textbf{32.5}\\
        \tabrow{4} & Navigate & TF & 32.6 & 18.4 & - & - & - & - & - & - & - & - & - & - & - & - & - & - \\
        \tabrow{5} & Navigate & SF & 31.9 & 16.9 & - & - & - & - & - & - & - & - & - & - & - & - & - & - \\
        \tabrow{6} & Navigate & MIX & \textbf{61.1} & \textbf{48.6} & - & - & - & - & - & - & - & - & - & - & - & - & - & - \\
        \tabrow{7} & Joint & MIX & 47.9 & 25.6 & \textbf{22.9} & \textbf{11.9} & 40.3 & \textbf{30.6} & \textbf{72.2} & \textbf{57.5} & \textbf{74.3} & \textbf{47.5} & 59.7 & \textbf{41.2} & \textbf{71.5} & \textbf{54.4} & \textbf{44.4} & 22.5\\
        \bottomrule
    \end{tabular}}
    \vspace{1pt}
    \caption{\textbf{Skill Pre-training Results.} Success Rates for Test-Seen and Test-Unseen at the 8 skills used in pre-training.}
    \label{tab:atomic_skills}
\end{table*}

\subsection{Skill Pre-training}
Learning a general policy to perform multiple tasks jointly is quite challenging. Moreover, the common benchmarks for interactive tasks (e.g., ALFRED~\cite{ALFRED20}) are typically small compared to passive, static tasks (e.g., ImageNet classification), which adds to the challenges of learning a generalizable model. One benefit of using a hierarchical policy is that the high-level policy can be disentangled from the skill sub-policies, which enables skill pre-training. In general, our definition of \emph{skill} is meaningful interactions with minimal sequences of primitive actions. 

Our skills span a range of activities such as navigation (e.g., $\langle\texttt{GoTo}, \texttt{X}\rangle$), interaction (e.g., $\langle\texttt{Open}, \texttt{X} \rangle$) and generating an answer (e.g., $\langle \texttt{Answer}, \texttt{None} \rangle$). The full list of skills, except the VQA skill, is shown in the header of Table~\ref{tab:atomic_skills}. 
For interaction, we assume the agent is already close to the target thus requiring minimum primitive navigation.  
During pre-training, we put an agent into an environment and task the agent to complete atomic skills (e.g., \texttt{GoTo Apple}, \texttt{Open Fridge}, etc). We continuously sample plausible skill-object pairs and train the agent using the losses defined below. As an agent interacts with a scene, objects get pushed and moved around. As a result, we need to periodically reset and shuffle the environment after a fixed number of episodes. 

We train the model with a combination of teacher forcing (TF), student forcing (SF) and Proximal Policy Optimization (PPO) \cite{Schulman2017ProximalPO} algorithms.  
For TF and SF, the expert trajectories can be obtained by the shortest path trajectory, which is obtained using a planner that has access to the full state of the environment. The loss for $\pi^{\texttt{nav/act}}$ (for TF and SF) is defined as:
\begin{equation}
\begin{split}
    \mathcal{L}^{\texttt{nav/act}} = \frac{1}{N} \sum_{t=1}^N [& \mathcal{L}_c(a_t, a_t^*) + \mathcal{L}_c(\bar{p_t}, \bar{p_t}^*) +\\
    & \mathcal{L}_g  + L_{\texttt{focal}}^{\texttt{aux}} + L_1^{\texttt{aux}}],
\end{split}
\end{equation}
where $N$ is the number of steps, $\mathcal{L}_c$ is the cross-entropy loss, $\mathcal{L}_g$ is a weighted Gaussian log-likelihood loss for continuous policy gradient, and $a$ and $\bar{p}$ denote the primitive action and discretized interaction point. $a_t^*$ and $\bar{p}^*_t$ are the expert action at step $t$. Motivated by \cite{zhou2019objects}, we also add two auxiliary losses: $L_{\texttt{focal}}^{\texttt{aux}}$ -- penalty-reduced pixelwise logistic regression with focal loss and $L_1^{\texttt{aux}}$ -- $L_1$ loss for offset prediction over all the visible objects. More details about the loss functions are given in the appendix. 

Training with PPO is challenging even for atomic skills, given the large space of actions and sparse reward setting. Hence, in addition to the goal success reward, we add a few auxiliary rewards to help the agent learn correct actions. More specifically, the reward vector is defined as $r = [r_{\texttt{success}}, r_{\texttt{visible}}, r_{\texttt{act}}, r_{\texttt{point}}]$ and the corresponding weights are defined in the appendix. In the reward vector, $r_{\texttt{success}} = 1$ if the sub-goal has achieved. $r_{\texttt{visible}}=1$ if the target object is visible. $r_{\texttt{act}} = 1$ if the agent takes the correct primitive action (compared to the expert planner). $r_{\texttt{point}}$ is a 2-d Normal distribution where the mean is the ground truth point on the object. The agent can obtain partial rewards even if it is not successful in accomplishing the sub-goal. Note that $r_{\texttt{point}}$ will be equal to zero for skills that do not require object interaction. We find that these auxiliary rewards greatly benfit training with PPO.

\subsection{Joint Multi-Task Training}

The pre-training stage trains sub-policies to perform atomic skills in the environment but not how to communicate with the high-level policy to accomplish the tasks. We now train the high level policy and finetune the sub-policies jointly for multiple high-level tasks. The overall loss for the high-level policy $\pi_\theta$ and sub-policies $\pi_{\theta^s}$ is defined as:
\begin{equation}
    \mathcal{L} = \frac{1}{N} \sum_{t=1}^N\left[ \mathcal{L}_c(g_{s_t}, g_{s_t}^*) + \mathcal{L}_c(g_{o_t}, g_{o_t}^*) + \mathbbm{1}_{g_{s_t}} \mathcal{L}^{g_{s_t}}  \right],
\end{equation}
where $\mathbbm{1}_{g_{s_t}}$ is an indicator function and $\mathcal{L}^{g_{s_t}}$ is the corresponding sub-policy loss for skill $g_{s_t}$. 

We use a \textit{recovery planner} to supervise the learning process of the high-level policy. This planner is defined as a dynamic planner for the high-level policy that can guide the agent to recover from any previous wrong actions.
For example, for a given sub-goal ``pick up apple'', the agent might pick up a nearby ``orange" by mistake. The agent cannot pick up the apple unless the agent 
drops the orange first. During training, we monitor the expert plan for the sub-goal that is being executed and the actual action the agent took. If the agent performs a wrong interactive action, the recovery planner inserts a new sub-goal to reverse the effect of the previous wrong action.
The ability to recover from failed actions is essential for high-level tasks, especially for long-horizon tasks. The use of a recovery planner is critical when training with student forcing.

During multi-task training, we randomly sample the episodes in proportion to the original task distribution and update the high-level policy and the corresponding sub-policies simultaneously. See appendix for more details on training the high level policy.

\section{Experiments}
\label{sec:experiments}


\begin{table*}\footnotesize
\setlength\tabcolsep{6pt}
    \center
    \begin{tabular}{ccl|rrrrrrrrrr}
        \toprule
        Interaction & Pretrain & Train & \multicolumn{2}{c}{\taskshort} & \multicolumn{2}{c}{\tasklong} & \multicolumn{2}{c}{\taskiqa} & \multicolumn{2}{c}{\taskexp} & \multicolumn{2}{c}{Averages}\\
        & & & \emph{{Seen}} & \emph{{Unseen}} & \emph{{Seen}} & \emph{{Unseen}} & \emph{{Seen}} & \emph{{Unseen}} & \emph{{Seen}} & \emph{{Unseen}} & \emph{{Seen}} & \emph{{Unseen}}\\
        \midrule
        \multicolumn{13}{l}{\colorbox{rowgray}{Model: \net (\netsmall)}}\\
        \multirow{2}{*}{\textbf{Standard}} & \multirow{2}{*}{No} & \tabrowb{1} Single & 75.9 & 7.1 & 5.1 & 0.1 & 45.8 & 12.8 & 21.6 & 8.1 & 37.0	& 7.0\\
        & & \tabrowb{2} Multi & 42.4 & 3.0 & 2.8 & 0.0 & 14.6 & 14.4 & 0.8 & 0.0 & 15.1 & 4.3\\
        \cmidrule(r){2-13}
        Uses & \multirow{2}{*}{Yes} & \tabrowb{3} Single & 77.4 & 25.7 & \textbf{10.5} & 0.8 & 46.2 & 18.0 & 16.9 & 5.2 & 37.7 & 12.4\\
        Detector & & \tabrowb{4} Multi &  72.3 & 29.8 & 9.4 & \textbf{1.3} & 55.0 & 20.3 & \textbf{31.2} & \textbf{13.6} & \textbf{41.9} & 16.2\\        
        \midrule
        
        \multirow{2}{*}{\textbf{Hard}} & \multirow{2}{*}{No} & \tabrowb{5} Single & \textbf{83.1} & 2.1 & 1.1 & 0.1 & 45.3 & 12.9 & 12.7 & 4.8 & 35.6 & 5.0\\
        
        &  & \tabrowb{6} Multi & 38.4 & 1.3 & 0.1 & 0.1 & 42.5 & 14.4 & 0.4 & 0.0 & 20.3 & 4.0\\
        \cmidrule(r){2-13}
        Predicts & \multirow{2}{*}{Yes} & \tabrowb{7} Single &  80.8 & 20.0 & 6.0 & 0.3 & 46.4 & 17.8 & 14.1 & 5.7 & 36.8 & 10.9\\
        Point & & \tabrowb{8} Multi & 71.7 & \textbf{42.7} & 5.4 & 0.6 & \textbf{55.4} & \textbf{22.3} & 26.7 & {11.4} & {39.8} & \textbf{19.3}\\
        \hline
        \hline
        \multicolumn{13}{l}{\colorbox{rowgray}{Model: Flat}}\\
        \multirow{2}{*}{\textbf{Hard}} & \multirow{2}{*}{No} & \tabrowb{9} Single & 1.0 & 0.0 & 0.7 & 0.0 & 14.2 & 9.2 & 1.9 & 0.8 & 4.5 & 2.5\\
        & & \tabrowb{10} Multi & 22.5 & 0.5 & 0.3 & 0.0 & 11.9 & 6.7 & 0.4 & 0.0 & 8.8 & 1.8\\
        \bottomrule
    \end{tabular}

    \vspace{1pt}
    \caption{\textbf{Multi-task Results.} Success Rates for Test-Seen and Test-Unseen for the 4 tasks. We evaluate 2 models, in single and multi-task settings in 2 interaction modes and also compare \asc pre-training to no-pretraining. Average metrics across all 4 tasks are also reported.}
    \label{tab:task_eval}
\end{table*}

\textbf{Dataset.} We train and evaluate our embodied agent within the \thor environment. We use 112 scenes across 4 scene types (kitchen, living rooms, bedrooms and bathrooms), and train our agent to complete four types of tasks -- (1) \textit{short-horizon} instruction following (\taskshort), (2) \textit{long-horizon} instruction following (\tasklong), (3) \textit{interactive question answering} (\taskiqa), and (4) \textit{exploratory interaction} (\taskexp). All tasks require the agent to interact with objects in the environment. The variety of scenes (112), task types (4) and target object categories (110) make the dataset very challenging.

The dataset contains 42,037 episodes split into 33,487/1,391/1,358/3,217/2,584 for Training/Val-Seen/Val-Unseen/Test-Seen/Test-Unseen, respectively. At the start of each episode, the agent's starting location is randomized and objects in a scene are automatically placed at random locations following a set of commonsense rules provided by \thor. Hence, no two episodes share the same configuration of objects. The Val-Seen and Test-Seen episodes are performed in the same scenes as Train (hence the suffix \emph{Seen}), but the configurations of the agent and objects are novel. In the Unseen splits, both the environments and object configurations are new to the agent. Please refer to the appendix for dataset and tasks details -- some crucial ones are presented below.

\noindent\textbf{SHIF.} These tasks are decomposed into a sequence of skills.~E.g., ``clean the apple" is decomposed into 
$\langle\texttt{GoTo}, \texttt{Sink}\rangle$--$\langle\texttt{Put}, \texttt{Sink}\rangle$--$\langle\texttt{ToggleOn}, \texttt{Faucet}\rangle$--$\langle\texttt{ToggleOff}, \texttt{Faucet}\rangle$--$\langle\texttt{Pickup}, \texttt{Apple}\rangle$.~We initialize the episodes with fulfilled pre-conditions i.e. apple is already in hand in this example.\\
\textbf{LHIF.} We follow the setting of ALFRED~\cite{ALFRED20} which consists of 7 different task types parameterized by 84 object classes. Crucially, we differ from ALFRED in that we only use the \emph{goal} instruction and not the step-by-step details. This results in a significantly harder setting as also noted in~\cite{ALFRED20}.\\
\textbf{IQA.} For \taskiqa, we follow the setting of the IQA dataset \cite{gordon2018iqa}. Given a question (e.g., ``How many bottles are in the fridge?"), the agent needs to navigate to the fridge and open it to answer the question. There are three different question types -- state questions, existence questions and counting questions, and the answer vocabulary is \{Yes, No, 0, 1, 2, 3\}. For each question type, we sample episodes with different scene configurations to make sure there is no bias that can be trivially exploited.\\
\textbf{EXIN.} The \taskexp task requires the agent to navigate to a target object, often very far away, and interact with it (e.g., pickup the apple, close the fridge, etc). To create the episode, we randomly initialize the agent in the room and randomly sample a target object and skill. 
For object-state-change skills, we ensure that the target object's state differs from its goal state.
For skills that require additional objects as pre-conditions (e.g., ``slice the apple'' requires the knife in agent's hand), the pre-condition is fulfilled at the beginning of the episode. 

\textbf{Evaluating Pre-training.}
Table~\ref{tab:atomic_skills} details performance at the 8 navigation and interaction atomic skills used for pre-training. We compare sub-policies trained with teacher forcing (TF), student forcing (SF) and a progression of TF $\rightarrow$ SF $\rightarrow$ reinforcement learning with PPO (MIX). We also compare training a joint sub-policy for all 8 skills vs 2 sub-policies, one for navigation  (Navigate) and one for interaction (Interact) skills. We report the Success Rate on the test-seen and test-unseen sets.

We observe that:
(1) For many skills, TF $\rightarrow$ SF $\rightarrow$ PPO provides gains over TF and SF; in some cases the gains are very large (18.4 to 48.6 for \texttt{GoTo} -- Row\texttt{\textcolor{blue}{4}} vs Row\texttt{\textcolor{blue}{6}}). (2) Joint training improves over Interact (Row\texttt{\textcolor{blue}{3}} vs Row\texttt{\textcolor{blue}{7}}) but it is much worse than Navigate (Row\texttt{\textcolor{blue}{6}} vs Row\texttt{\textcolor{blue}{7}}) -- likely because properties of the navigation skill (such as length) are vastly different from others. (3) Skills that require interacting with very small objects (\texttt{PickUp} often picks up objects such as a pencil or knife) tend to be very challenging, since the target object is difficult to pin point. (4) The skill Success rates in Seen and Unseen rooms are encouraging, given the challenging environment, and useful for downstream tasks. The 9th skill (providing an answer), not shown in Table~\ref{tab:atomic_skills}, uses the last image from the expert policy. Here, we obtain an accuracy of 76.0 on Seen and 52.5 on Unseen scenes.

%

\begin{figure*}[th!]
    \centering
    \includegraphics[width=1.05\textwidth]{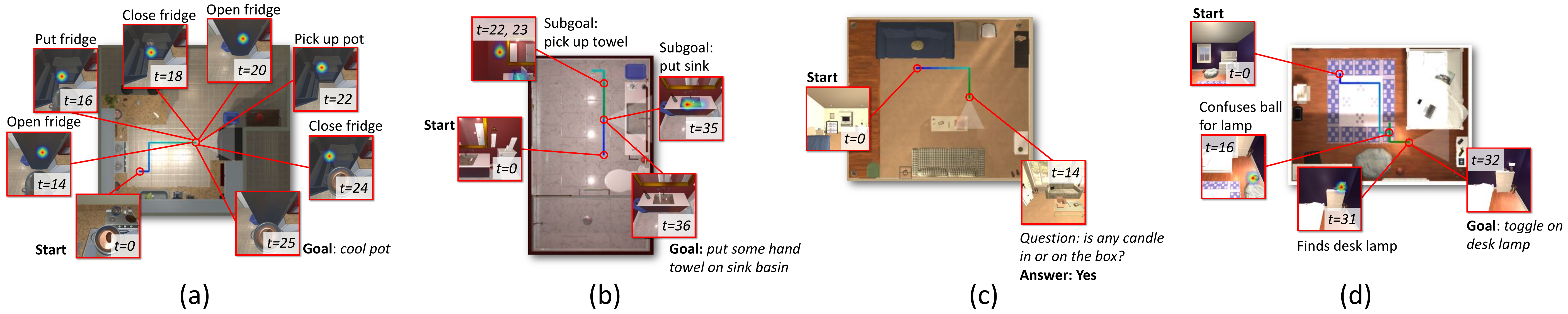}
    \caption{\textbf{Qualitative Results.} \netsmall trajectories for validation episodes in Hard setting allow us to better interpret the observations and actions of the agent. (a) \taskshort, with a large amount of interactions in a \emph{kitchen}, (b) \tasklong with multiple interaction and navigation sub-goals in a \emph{bathroom}, (c) \taskiqa for a binary question in a \emph{living room}, and (d) \taskexp with exploration, incorrect identification of the target object and posterior correction in a \emph{bedroom}.}
    \label{fig:qualitative}
\end{figure*}

\begin{table*}[tp!]\footnotesize
\setlength\tabcolsep{4pt}
    \center
    \resizebox{\textwidth}{!}{
    \begin{tabular}{cccccccccccccc}
        \toprule
        Student & Random & Recovery & Pre- & \multicolumn{2}{c}{\taskshort} & \multicolumn{2}{c}{\tasklong} & \multicolumn{2}{c}{\taskiqa} & \multicolumn{2}{c}{\taskexp} & 
        \multicolumn{2}{c}{Averages}\\
        Forcing & init & Planner & training & seen & unseen & seen & unseen & seen & unseen & seen & unseen & seen & unseen \\
         \midrule
         - & \ding{51} & \ding{51} & \ding{51} & 69.13 & 24.53 & 4.09 & 0.29 & 54.10 & 32.10 & 21.66 & 10.96 & 37.24 & 16.97\\
         \ding{51} & - & \ding{51} & \ding{51} & \textbf{72.90} & 22.13 & 4.58 & 0.43 & 54.40 & \textbf{26.03} & 26.45 & \textbf{12.85} & 39.58 & 15.36\\
         \ding{51} & \ding{51} & - & \ding{51} & 64.10 & 22.77 & 3.81 & 0.64 & 49.43 & 18.33 & 21.83 & 11.43 & 34.79 & 13.29\\
         \ding{51} & \ding{51} & \ding{51} & - & 38.40 & 1.31 & 0.10 & 0.14 & 42.47 & 14.43 & 0.40 & 0.00 & 20.34 & 3.97\\
         \ding{51} & \ding{51} & \ding{51} &  \ding{51} & 71.70 & \textbf{42.70} & \textbf{5.43} & \textbf{0.64} & \textbf{55.43} & 22.33 & \textbf{26.66} & 11.42 & \textbf{39.80} & \textbf{19.27}\\
        \bottomrule
    \end{tabular}
    }
    \vspace{2pt}
    \caption{\textbf{Ablation Study} for \netsmall on the Test sets.} 
    \label{tab:ablation}
\end{table*}

\textbf{Evaluating Multi-task Training.}
Table~\ref{tab:task_eval} details Success Rates in the Test scenes of several single and multi-task agents at the 4 tasks and also reports averages across all 4, separated by the Seen and Unseen splits. We report performance for 2 models: our proposed \netsmall and a baseline, \netbase -- which is not hierarchical, and identical to our interactive sub-policy except the model takes task embedding instead of sub-goal embedding as input. We train both models in a single and multi-task setup. Further, \netsmall is trained with and without \asc pre-training in the Standard and Hard interactive settings. When trained without \asc, the sub-policy is initialized from scratch.

Given the large set of results, we refer to the Averages columns in the text below but encourage the reader to look at all columns in Table~\ref{tab:task_eval}. We observe that: (1) Pre-training the agent with \asc provides very large gains across all 4 tasks when compared to no pre-training. For the multi-task training setup, these large gains are seen for both Standard (Row\texttt{\textcolor{blue}{2}} vs Row\texttt{\textcolor{blue}{4}}) and Hard (Row\texttt{\textcolor{blue}{6}} vs Row\texttt{\textcolor{blue}{8}}) settings. The Average columns show that the improvements are on the order of 2x for Seen and 4x for Unseen scenes. (2) \asc enables us to train effective multi-task agents. In the absence of pre-training, multi-task training results in a drop over single task training (Row\texttt{\textcolor{blue}{6}} vs Row\texttt{\textcolor{blue}{5}}), but with \asc, we see gains in going to multi-task (Row\texttt{\textcolor{blue}{8}} vs Row\texttt{\textcolor{blue}{7}}). (3) Our agent performs comparably well in the Hard setting (Row\texttt{\textcolor{blue}{4}} vs Row\texttt{\textcolor{blue}{8}}) indicating that the network is effective at localizing points on target objects. (4) \netsmall outperforms \netbase by huge margins (Row\texttt{\textcolor{blue}{8}} vs Row\texttt{\textcolor{blue}{10}}).

\textbf{Ablation Study.}
Table~\ref{tab:ablation} presents an ablation study. We ablate the effects of student forcing, random initialization of scenes, usage of the recovery planner and pre-training with \asc. As seen, removing each of these components provides a drop in the Seen and Unseen success rates on average across all four tasks. The largest drop is observed if pre-training is switched off indicating the immense benefit of pre-training. Random initialization is expectedly useful for generalization.


\textbf{Interpretability.} Figure~\ref{fig:qualitative} shows trajectories for Val-Seen episodes in the Hard interaction setting for each of the four task types. As observed, our hierarchical agent is able to solve them effectively. Importantly, our method is more interpretable than past approaches that directly output an action based on the current observation and the language specification of the task (e.g., \cite{pashevich2021episodic}). At each time step, one can observe the sub-goals and pixel locations output by the high level policy. The sub-goals allow us to interpret the progress of the agent along its episode and its current sub-goal of interest, and the pixel heatmaps and object types allow us to interpret which object the agent is presently interested in interacting with, and where it thinks the object resides in the scene. For instance, in (b) one can notice sub-goals like \texttt{Pick up towel} and \texttt{Put sink} that are correctly executed. Also notice an error in (d) where the high level policy confuses a ball by the desk lamp, but the agent eventually recovers to find the correct desk lamp and then switch it on.

\section{Conclusion}
Solving Embodied AI tasks requires tackling unique challenges due to the long-horizon nature of the tasks, partial observability of the states and high-dimensional inputs such as images. While there has been significant progress in multi-task training in vision and NLP, less progress has been seen in Embodied AI, where models usually target an individual task. As a step towards multi-task training for Embodied AI, we propose \asc, a method that proves effective in training multiple tasks jointly. A key element of this approach is a pre-training strategy and a training regime for handling multiple embodied tasks. Our experimental evaluations show that our multi-task training approach provides better results compared to training each task individually, while the amortized amount of data for each task is significantly lower.


\textbf{Limitations}: The proposed approach has certain limitations. We discuss a few important ones here. First, the pre-training strategy relies on supervision from the environment. While the use of simulated environments makes this plausible, pre-training skills using only self supervision is an interesting area of research that we will address in future work. Second, the set of skills are manually defined. Automatic learning of the required skills is an interesting direction to explore. Finally, this work abstracts away a lot of challenges involved in physical robot interactions. Given the numerous challenges, even in simulation, transfer to the physical world will be considered in future work. Having said that, we have tried to minimize the assumptions that are only valid in simulation.


\textbf{Negative societal impact}: The scope of our contributions do not have a direct negative societal impact. Research in the Embodied AI domain might lead to creating robots that can be used for malicious applications. While important to consider, we do not posit any imminent concern given the numerous challenges that remain in our quest to build autonomous and intelligent agents.

{\small
\bibliographystyle{bib/ieee_fullname}
\bibliography{bib/main.bib}
}

\newpage
\appendix
\section*{Appendix}
\section{Implementation Details}
Here we provide the implementation details of our full model. There are two stages:

\textbf{Stage 1: Skill Pre-training.} For interaction, navigation and question answering sub-policies, we use a ResNet18 model pre-trained on ImageNet as the backbone. The first two residual blocks in the backbone are fixed during
training. We periodically reset and randomly initialize the objects in the room every 10 rollout steps. We use Adam with learning rate of $3\cdot10^{-4}$., and train with teacher forcing (20M steps), student forcing (20M steps) and proximal policy optimization (60M steps). 
For student forcing, the agent alternates between choosing actions from ground-truth planner or from the current learned policy with some probability $\epsilon$. We use a linear decay schedule $(\epsilon=1.0 \rightarrow 0.0)$ in our experiment.
The corresponding weight for the reward vector $r = [r_{\texttt{success}}, r_{\texttt{visible}}, r_{\texttt{act}}, r_{\texttt{point}}]$ is $[20.0, 1.0, 1.0, 0.5]$.
Our model can be trained with 8 Titan X GPU with 72 processes in 5 days.

\textbf{Stage 2: Multi-Task Training.}
For multi-task training, we initialize the sub-policies with the pre-trained model. We use the same ResNet18 model pre-trained on ImageNet as the backbone, and all residual blocks in the backbone are fixed during training. 
We use Adam with learning rate of 3e-4 to train the high-level policy and 3e-5 to finetune the sub-policies.
For long horizon tasks such as \tasklong, it is very hard to successfully accomplish the task by random exploration of the environment. Thus we only train with teacher forcing and student forcing and do not use PPO.  For all models, we train with teacher forcing (10M steps) and student forcing (10M steps). 
Similar to skill pre-training, we use a linear decay schedule $(\epsilon=1.0 \rightarrow 0.6)$ in our experiment.
For multi-task training, our model can be trained with 8 Titan X GPU with 40 processes in 2 days. 

\section{Action Space} 
Here we provide the details of our action space:

\textbf{High-level policy.} Our high-level policy predicts the skills and target object type for the sub-policies. There are 10 skills including \texttt{<End>} (indicating the end of the execution) and 110 target objects. The skills are \texttt{<Goto>}, \texttt{<Pickup>}, \texttt{<Put>}, \texttt{<ToogleOn>}, \texttt{<ToogleOff>}, \texttt{<Open>}, \texttt{<Close>}, \texttt{<Slice>}, \texttt{<Answer>} and \texttt{<End>}. Since \texttt{<Answer>} and \texttt{<End>} do not take any target object, there are 8 $\times$ 110 + 2 = 882 possible choices at each time step for the high-level policy.

\textbf{Navigation sub-policy.} The agents navigate through the environment via 6 different actions \texttt{MoveAhead}, \texttt{RotateLeft}, \texttt{RotateRight}, \texttt{LookUp},
\texttt{LookDown} and \texttt{Done}.

\textbf{Interaction sub-policy.} For interaction sub-policy, we assume the agent is close to the target, thus needs to navigate and perform the interaction action. Thus the action space is \texttt{MoveAhead}, \texttt{RotateLeft}, \texttt{RotateRight}, \texttt{LookUp},
\texttt{LookDown}, \texttt{OpenObject}, \texttt{CloseObject}, \texttt{PickupObject}, \texttt{PutObject}, \texttt{ToggleObjectOn}, \texttt{ToggleObjectOff}, \texttt{SliceObject} and \texttt{Done}. For interactive actions, the interaction sub-policy also needs to predict an interaction point on the image plane which is specified by a discrete location on a grid and a continuous offset from the point on the grid. We consider an $8\times 8$ grid, the continuous offset is sampled from a 2-d multivariate normal distribution. 

\textbf{Question answering sub-policy.} The action space for question answering sub-policy is \texttt{Yes}, \texttt{No}, \texttt{0}, \texttt{1}, \texttt{2} and \texttt{3}.

\begin{table*}[ht]\scriptsize
\setlength\tabcolsep{2pt}
    \center
    \resizebox{\textwidth}{!}{
    \begin{tabular}{ccl|rrrrrrrr|rrrrrrrr}
        \toprule
        Interaction & Pretrain & Train & \multicolumn{2}{c}{Heat} & \multicolumn{2}{c}{Clean} & \multicolumn{2}{c}{Cool} &  \multicolumn{2}{c}{Averages} & \multicolumn{2}{c}{Existing} & 
        \multicolumn{2}{c}{Counting} & 
        \multicolumn{2}{c}{State} & 
        \multicolumn{2}{c}{Averages}\\
        & & & \emph{\tiny{Seen}} & \emph{\tiny{Unseen}} & \emph{\tiny{Seen}} & \emph{\tiny{Unseen}} & \emph{\tiny{Seen}} & \emph{\tiny{Unseen}} & \emph{\tiny{Seen}} & \emph{\tiny{Unseen}} &
        \emph{\tiny{Seen}} & \emph{\tiny{Unseen}} & 
        \emph{\tiny{Seen}} & \emph{\tiny{Unseen}} & 
        \emph{\tiny{Seen}} & \emph{\tiny{Unseen}} & 
        \emph{\tiny{Seen}} & \emph{\tiny{Unseen}} \\
        \midrule
        \multicolumn{19}{l}{\colorbox{rowgray}{Model: \net (\netsmall)}}\\
        \multirow{2}{*}{\textbf{Standard}} & \multirow{2}{*}{No} & \tabrowb{1} Single & 88.0 & 2.3 & 54.0 & 6.2 & 85.7 & 12.7 & 75.9 & 7.1 & 55.2 & 18.5 & 36.2 & 8.8 & 46.0 & 11.1 & 45.8 & 12.8 \\
        & & \tabrowb{2} Multi & 41.0 & 4.7 & 29.9 & 1.2 & 56.2 & 3.2 & 42.4 & 3.0 & 17.5 & 16.3 & 10.5 & 10.2 & 15.9 & 16.7 & 14.6 & 14.4 \\
        \cmidrule(r){2-19}
        Uses & \multirow{2}{*}{Yes} & \tabrowb{3} Single & 87.5 & 23.3 & 54.7 & 17.3 & 90.0 & 36.5 & 77.4 & 25.7 & 54.0  & 20.0 & 32.5 & 11.0 & 52.1 & 22.9 & 46.2 & 18.0 \\
        Detector & & \tabrowb{4} Multi &  81.0 & 25.6 & 55.5 & 32.1 & 80.4 & 31.7 & 72.3 & 29.8 & 66.5 & 25.0 & 39.3 & 12.0 & 59.3 & 23.8 & 55.0 & 20.3 \\        
        \midrule
        
        \multirow{2}{*}{\textbf{Hard}} & \multirow{2}{*}{No} & \tabrowb{5} Single & 94.0 & 0.0 & 59.9 & 0.0 & 95.5 & 6.4 & 83.1 & 2.1 & 54.2 & 18.8 & 36.0 & 8.8 & 45.8 & 11.1 & 45.3 & 12.9 \\
        
        &  & \tabrowb{6} Multi & 40.0 & 2.3 & 19.0 & 0.0 & 56.2 & 1.6 & 38.4 & 1.3 & 17.5 & 16.3 & 10.5 & 10.2 & 15.9 & 16.7 & 14.6 & 14.4\\
        \cmidrule(r){2-19}
        Predicts & \multirow{2}{*}{Yes} & \tabrowb{7} Single &  90.0 & 23.3 & 56.9 & 4.9 & 95.5 & 31.7 & 80.8 & 20.0 & 54.2 & 20.0 & 32.5 & 11.0 & 52.7 & 22.2 & 46.5 & 17.7\\
        Point & & \tabrowb{8} Multi & 84.0 & 23.3 & 40.9 & 52.4 & 90.2 & 52.4 & 71.7 & 42.7 & 67.5 & 25.5 & 39.5 & 11.7 & 59.3 & 23.8 & 55.4 & 20.3 \\
        \hline
        \hline
        \multicolumn{19}{l}{\colorbox{rowgray}{Model: Flat}} \\
        \multirow{2}{*}{\textbf{Hard}} & \multirow{2}{*}{No} & \tabrowb{9} Single & 0.0 & 0.0 & 0.0 & 0.0 & 3.1 & 0.0 & 1.0 & 0.0 & 14.2 & 8.3 & 6.8 & 2.3 & 21.6 & 17.1 & 14.2 & 9.2\\
        & & \tabrowb{10} Multi & 21.0 & 0.0 & 23.4 & 0.0 & 23.2 & 1.59 & 22.5 & 0.5 & 13.8 & 5.5 & 9.3 & 5.5 & 12.5& 9.2 & 11.9 & 6.7 \\
        \bottomrule
    \end{tabular}
    }
    \vspace{1pt}
    \caption{Short Horizon Instruction Following (\taskshort) and Interactive QA (\taskiqa) results.}
    \label{tab:table_detail_1}
    \vspace{-4mm}
\end{table*}

\begin{table*}[h]\scriptsize
\setlength\tabcolsep{2pt}
    \center
    \resizebox{\textwidth}{!}{
    \begin{tabular}{ccl|rrrrrrrrrrrrrrrr}
        \toprule
        Interaction & Pretrain & Train & \multicolumn{2}{c}{Pick} & \multicolumn{2}{c}{Stack} & \multicolumn{2}{c}{Pick Two} & \multicolumn{2}{c}{Clean} &
        \multicolumn{2}{c}{Heat} &
        \multicolumn{2}{c}{Cool} &
        \multicolumn{2}{c}{Examine} &
        \multicolumn{2}{c}{Averages}\\
         &  &  & \multicolumn{2}{c}{\& Place} & \multicolumn{2}{c}{\& Place} & \multicolumn{2}{c}{\& Place} & \multicolumn{2}{c}{\& Place} &
        \multicolumn{2}{c}{\& Place} &
        \multicolumn{2}{c}{\& Place} &
        \multicolumn{2}{c}{in Light} &
        \multicolumn{2}{c}{}\\
        & & & \emph{\tiny{Seen}} & \emph{\tiny{Unseen}} & \emph{\tiny{Seen}} & \emph{\tiny{Unseen}} & \emph{\tiny{Seen}} & \emph{\tiny{Unseen}} & \emph{\tiny{Seen}} & \emph{\tiny{Unseen}} & \emph{\tiny{Seen}} & \emph{\tiny{Unseen}} & 
        \emph{\tiny{Seen}} & \emph{\tiny{Unseen}} & 
        \emph{\tiny{Seen}} & \emph{\tiny{Unseen}} & 
        \emph{\tiny{Seen}} & \emph{\tiny{Unseen}}\\
        \midrule
        \multicolumn{19}{l}{\colorbox{rowgray}{Model: \net (\netsmall)}}\\
        \multirow{2}{*}{\textbf{Standard}} & \multirow{2}{*}{No} & \tabrowb{1} Single &  16.7 & 1.0 & 4.0 & 0.0 & 1.3 & 0.0 & 0.0 & 0.0 & 1.3 & 0.0 & 5.3 & 0.0 & 6.7 & 0.0 & 5.1 & 0.1\\
        & & \tabrowb{2} Multi & 10.0 & 0.0 & 0.7 & 0.0 & 0.7 & 0.0 & 0.0 & 0.0 & 1.3 & 0.0 & 2.7  & 0.0 & 4.0 & 0.0 & 2.8 & 0.0 \\
        \cmidrule(r){2-19}
        Uses & \multirow{2}{*}{Yes} & \tabrowb{3} Single & 22.0 & 1.0 & 8.7 & 0.0 & 6.7 & 0.0 & 2.0 & 0.0 & 6.0 & 0.0 & 9.3 & 0.0 & 18.7 & 4.4 & 10.5 & 0.8\\
        Detector & & \tabrowb{4} Multi & 22.7 & 5.0 & 10.7 & 0.0 & 4.0 & 0.0 & 2.7 & 0.0 & 2.0 & 1.4 & 6.7 & 1.0 & 17.3 & 1.5 & 9.4 & 1.3 \\        
        \midrule
        
        \multirow{2}{*}{\textbf{Hard}} & \multirow{2}{*}{No} & \tabrowb{5} Single & 4.0 & 1.0 & 2.7 & 0.0 & 0.0 & 0.0 & 0.0 & 0.0 & 0.0 & 0.0 & 0.0 & 0.0 & 1.3 & 0.0 & 1.1 & 0.1\\
        
        &  & \tabrowb{6} Multi & 0.7 & 1.0 & 0.0 & 0.0 & 0.0 & 0.0 & 0.0 & 0.0 & 0.0 & 0.0 & 0.0 & 0.0 & 0.0 & 0.0 & 0.1 & 0.1\\
        \cmidrule(r){2-19}
        Predicts & \multirow{2}{*}{Yes} & \tabrowb{7} Single & 12.7 & 2.0 & 10.0 & 0.0 & 1.3 & 0.0 & 4.0 & 0.0 & 1.3 & 0.0 & 4.7 & 0.0 & 8.0 & 0.0 & 6.0 & 0.3\\
        Point & & \tabrowb{8} Multi &  15.3 & 3.0 & 10.7 & 0.0 & 1.3 & 0.0 & 0.7 & 0.0 & 0.0 & 0.0 & 2.7 & 0.0 & 7.3 & 1.5 & 5.4 & 0.6\\
        \hline
        \hline
        \multicolumn{19}{l}{\colorbox{rowgray}{Model: Flat}}\\
        \multirow{2}{*}{\textbf{Hard}} & \multirow{2}{*}{No} & \tabrowb{9} Single & 2.0 & 0.0 & 2.7 & 0.0 & 0.0 & 0.0 & 0.0 & 0.0 & 0.0 & 0.0 & 0.0 & 0.0 & 0.0 & 0.0 & 0.7 & 0.0 \\
        & & \tabrowb{10} Multi & 2.0 & 0.0 & 0.0 & 0.0 & 0.0 & 0.0 & 0.0 & 0.0 & 0.0 & 0.0 & 0.0 & 0.0 & 0.0 & 0.0 & 0.3 & 0.0\\
        \bottomrule
    \end{tabular}
    }
    \vspace{1pt}
    \caption{Long Horizon Instruction Following (\tasklong) results.}
    \label{tab:table_detail_2}
    \vspace{-4mm}
\end{table*}

\begin{table*}[h]\scriptsize
\setlength\tabcolsep{2pt}
    \center
    \resizebox{\textwidth}{!}{
    \begin{tabular}{ccl|rrrrrrrrrrrrrrrrrr}
        \toprule
        Interaction & Pretrain & Train & \multicolumn{2}{c}{\texttt{Pickup}} & \multicolumn{2}{c}{\texttt{Put}} & \multicolumn{2}{c}{\texttt{ToggleOn}} & \multicolumn{2}{c}{\texttt{ToggleOff}} &
        \multicolumn{2}{c}{\texttt{Open}} &
        \multicolumn{2}{c}{\texttt{Close}} & 
        \multicolumn{2}{c}{\texttt{Slice}} & 
        \multicolumn{2}{c}{Averages}\\
        & & & \emph{\tiny{Seen}} & \emph{\tiny{Unseen}} & \emph{\tiny{Seen}} & \emph{\tiny{Unseen}} & \emph{\tiny{Seen}} & \emph{\tiny{Unseen}} & \emph{\tiny{Seen}} & \emph{\tiny{Unseen}} & \emph{\tiny{Seen}} & \emph{\tiny{Unseen}} &
        \emph{\tiny{Seen}} & \emph{\tiny{Unseen}} & 
        \emph{\tiny{Seen}} & \emph{\tiny{Unseen}} & 
        \emph{\tiny{Seen}} & \emph{\tiny{Unseen}} \\
        \midrule
        \multicolumn{19}{l}{\colorbox{rowgray}{Model: \net (\netsmall)}}\\
        \multirow{2}{*}{\textbf{Standard}} &  \multirow{2}{*}{No} & \tabrowb{1} Single & 11.0 & 6.7 & 0.0 & 0.0 & 11.1 & 6.7 & 38.9 & 20.0 & 22.0 & 0.0 & 27.8 & 6.67 & 40.0 & 16.7 & 21.6 & 8.1 \\
        & & \tabrowb{2} Multi & 0.0 & 0.0 & 0.0 & 0.0 & 0.0 & 0.0 & 2.8 & 0.0 & 0.0 & 0.0 & 2.8 & 0.0 & 0.0 & 0.0 & 0.8 & 0.0\\
        \cmidrule(r){2-19}
        Uses & \multirow{2}{*}{Yes} & \tabrowb{3} Single & 5.6 & 6.7 & 11.1 & 0.0 & 27.8 & 6.7 & 30.6 & 6.7 & 13.9 & 6.7 & 19.4 & 10.0 & 10.0 & 0.0 & 16.9 & 5.2\\
        Detector & & \tabrowb{4} Multi & 27.8 & 3.3 & 25.0 & 20.0 & 44.4 & 13.3 & 50.0 & 20.0 & 33.3 & 13.3 & 27.8 & 16.7 & 10.0 & 8.3 & 31.2 & 13.6 \\        
        \midrule
        
        \multirow{2}{*}{\textbf{Hard}} & \multirow{2}{*}{No} & \tabrowb{5} Single & 5.6 & 6.7 & 0.0 & 6.7 & 5.6 & 0.0 & 33.0 & 13.3 & 11.1 & 0.0 & 22.2 & 6.7 & 11.1 & 0.0 & 12.7 & 4.8\\
        
        &  & \tabrowb{6} Multi & 0.0 & 0.0 & 0.0 & 0.0 & 0.0 & 0.0 & 0.0 & 0.0 & 0.0 & 0.0 & 2.8 & 0.0 & 0.0 & 0.0 & 0.4 & 0.0 \\
        \cmidrule(r){2-19}
        Predicts & \multirow{2}{*}{Yes} & \tabrowb{7} Single & 5.6 & 0.0 & 11.1 & 0.0 & 19.4 & 6.7 & 22.2 & 13.3 & 13.9 & 6.7 & 16.7 & 13.3 & 10.0 & 0.0 & 14.1 & 5.7 \\
        Point & & \tabrowb{8} Multi & 13.9 & 6.7 & 22.2 & 10.0 & 33.3 & 13.3 & 38.9 & 16.7 & 33.3 & 10.0 & 25.0 & 23.3 & 20.0 & 0.0 & 26.7 & 11.4\\
        \hline
        \hline
        \multicolumn{19}{l}{\colorbox{rowgray}{Model: Flat}}\\
        \multirow{2}{*}{\textbf{Hard}} & \multirow{2}{*}{No} & \tabrowb{9} Single & 0.0 & 0.0 & 0.0 & 0.0 & 5.6 & 6.7 & 0.0 & 0.0 & 0.0 & 0.0 & 0.0 & 6.7 & 0.0 & 0.0 & 0.8 & 1.9\\
        & & \tabrowb{10} Multi & 0.0 & 0.0 & 0.0 & 0.0 & 2.8 & 0.0 & 0.0 & 0.0 & 0.0 & 0.0 & 2.8 & 0.0 & 0.0 & 0.0 & 0.8 & 0.0\\
        \bottomrule
    \end{tabular}
    }
    \vspace{1pt}
    \caption{Exploratory Interaction  (\taskexp) Results.}
    \label{tab:table_detail_3}
    \vspace{-4mm}
\end{table*}

\section{Auxiliary Loss Functions}

We add two auxiliary losses during skill pre-training: $L_{\texttt{focal}}^{\texttt{aux}}$ and $L_1^{\texttt{aux}}$. 
Following \cite{zhou2019objects}, we smooth out all ground truth object centers $Y \in [0, 1]^{\frac{W}{R}\times\frac{H}{R}\times C}$ using a Gaussian kernel $Y_{xyc}=\exp{(-\frac{((x-p_x)^2 + (y-p_y)^2)}{2\sigma^2_{pc}})}$, where $\sigma_{pc}$ is the standard deviation and depends on the object size. The training objective is penalty-reduced pixelwise logistic regression with focal loss:
\small
\begin{displaymath}
L_{\texttt{focal}}^{\texttt{aux}} = -\frac{1}{M}\sum_{xyc} \left\{ \begin{array}{ll}
(1-\hat{Y}_{xyc})^\alpha \log (\hat{Y}_{xyc}) & \textrm{if $Y_{xyc}=1$}\\
(1-Y_{xyc}^\beta) (\hat{Y}_{xyc})^\alpha \log(1-\hat{Y}_{xyc}) & \textrm{otherwise}
\end{array} \right.
\end{displaymath}

where $\alpha$ and $\beta$ are hyper-parameters of the focal loss ($\alpha=2$, $\beta=4$). $M$ is the number of object centers in image. The offset mean is shared across all classes and can be trained with an L1 loss:
\begin{equation}
    L_1^{\texttt{aux}} = \frac{1}{M}\sum_{p} \vert \mu_p - \mu_p^* \vert,
\end{equation}
where $\mu_p$ is the predicted mean of the offset and $\mu_p^*$ is the target offset. 

\section{Results of task sub-categories}
Table~2 shows the average performance. Here we show the detailed results for sub-categories of each task in Table~\ref{tab:table_detail_1}, Table~\ref{tab:table_detail_2} and Table~\ref{tab:table_detail_3}.   
\section{Dataset}

\begin{table*}[h]\footnotesize
\setlength\tabcolsep{25pt}
    \centering
    \begin{tabular}{ll}
        \toprule
        Task & Instruction Templates\\
        \midrule
        \multirow{1}{*}{\taskshort}  & clean \{obj\}; cool \{obj\}; heat \{obj\}\\
        \midrule
        \multirow{7}{*}{\tasklong} & put a \{obj\} in \{recep\}, put some \{obj\} on \{recep\}; \\
        & put a clean \{obj\} in \{recep\}, clean some \{obj\} and put it in \{recep\};\\
        &put a hot \{obj\} in \{recep\}, heat some \{obj\} and put it in \{recep\}; \\
        &put a cold \{obj\} in \{recep\}, cool some \{obj\} and put it in \{recep\}; \\ 
        & put two \{obj\} in \{recep\},find two \{obj\} and put them in \{recep\};\\
        & look at \{obj\} under the \{toggle\}, examine the \{obj\} with the \{toggle\};\\
        & put \{obj\} in a \{mrecep\} and then put them in \{recep\}, \\
        & put a \{mrecep\} of \{obj\} in \{recep\}, put \{obj\} \{mrecep\} in \{recep\};\\
        \midrule
        \multirow{4}{*}{\taskiqa} & is the \{obj\} \{state\}?; is any \{obj\} in or on the \{recep\}?, \\
        & does the \{recep\} contain or support at least one \{obj\}?; \\
        & how many \{obj\} are in or on the \{recep\}?, \\
        & count the number of \{obj\} in or on the \{recep\}; \\
        \midrule
        \taskexp & pick up \{obj\}; put \{obj\}; toggle on \{obj\}; toggle off \{obj\}; \\
        & open \{obj\}; close \{obj\}; slice \{obj\} \\
        \bottomrule
    \end{tabular}
    \vspace{2pt}
    \caption{Instruction templates for \taskshort, \tasklong, \taskiqa and \taskexp .}
    \label{tab:dataset_template}
\end{table*}

\begin{figure*}[h]
    \centering
    \newcommand\tasktypeheight{0.13\textwidth}
    \begin{tabular}{cc}
        \taskshort & \tasklong \\
        \multicolumn{1}{l}{\includegraphics[height=\tasktypeheight]{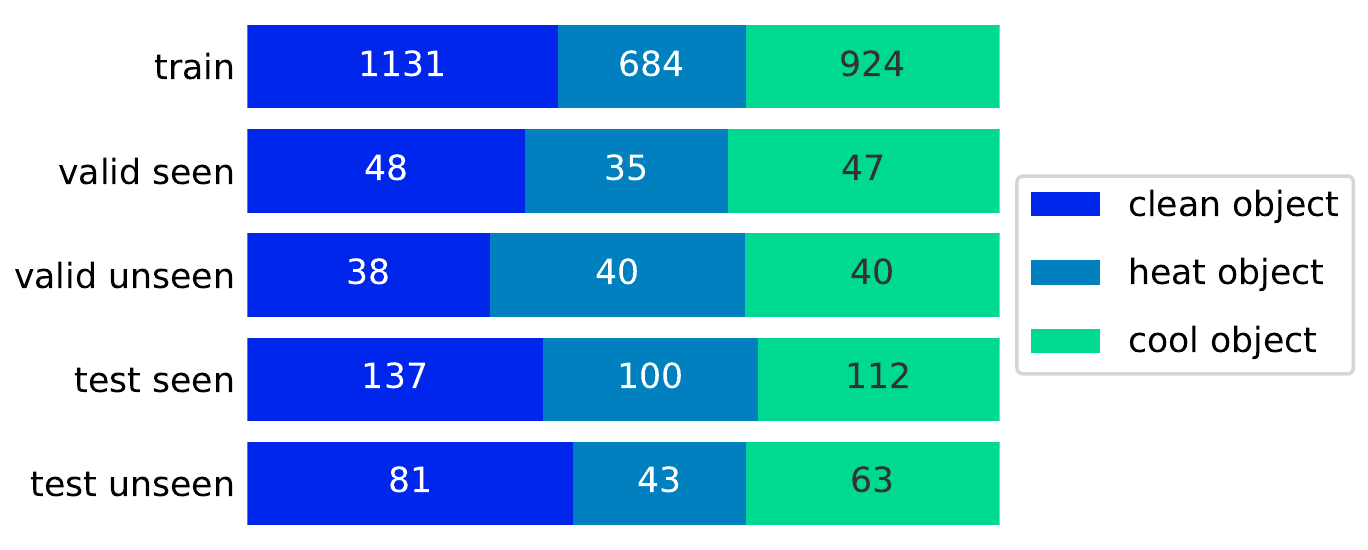}} &
        \multicolumn{1}{l}{\includegraphics[height=\tasktypeheight]{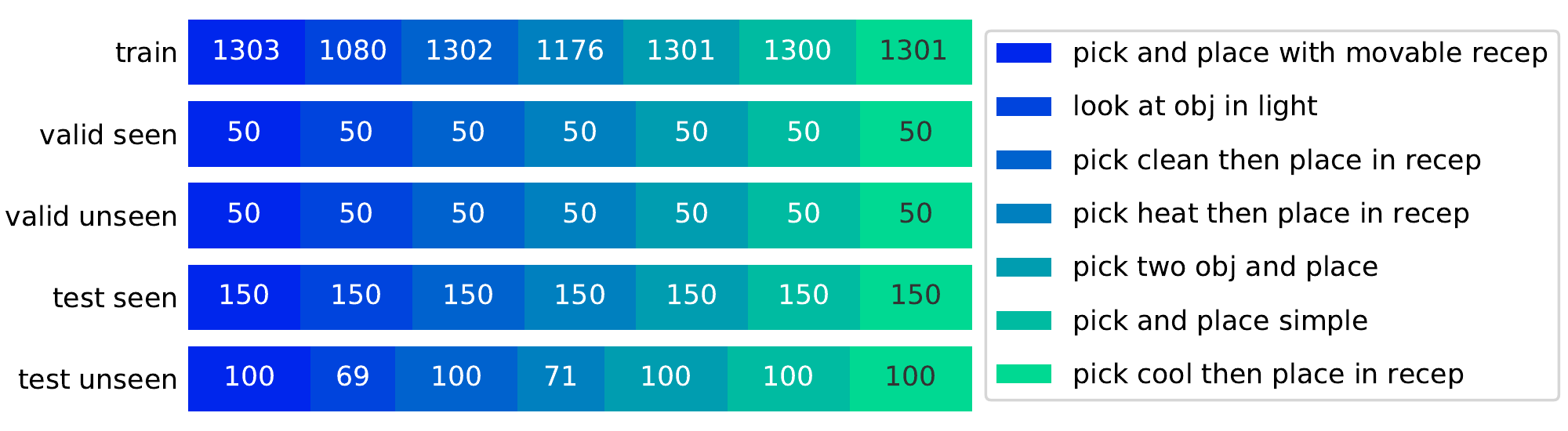}} \\
        \taskiqa & \taskexp \\
        \multicolumn{1}{l}{\includegraphics[height=\tasktypeheight]{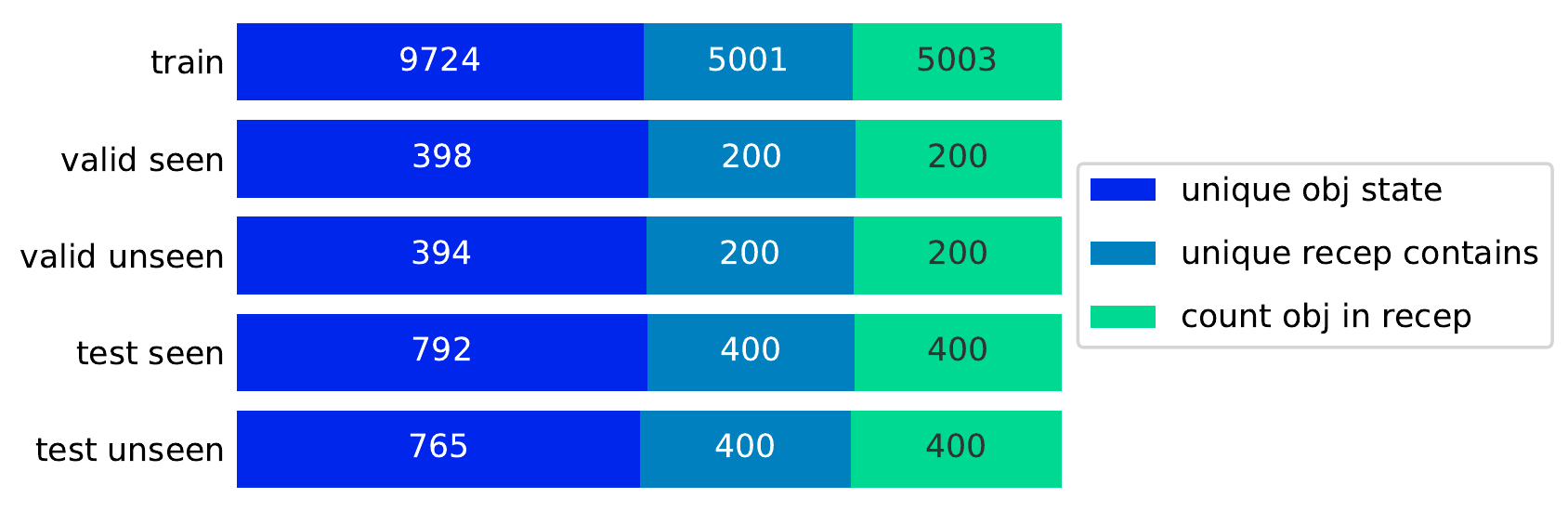}} &
        \multicolumn{1}{l}{\includegraphics[height=\tasktypeheight]{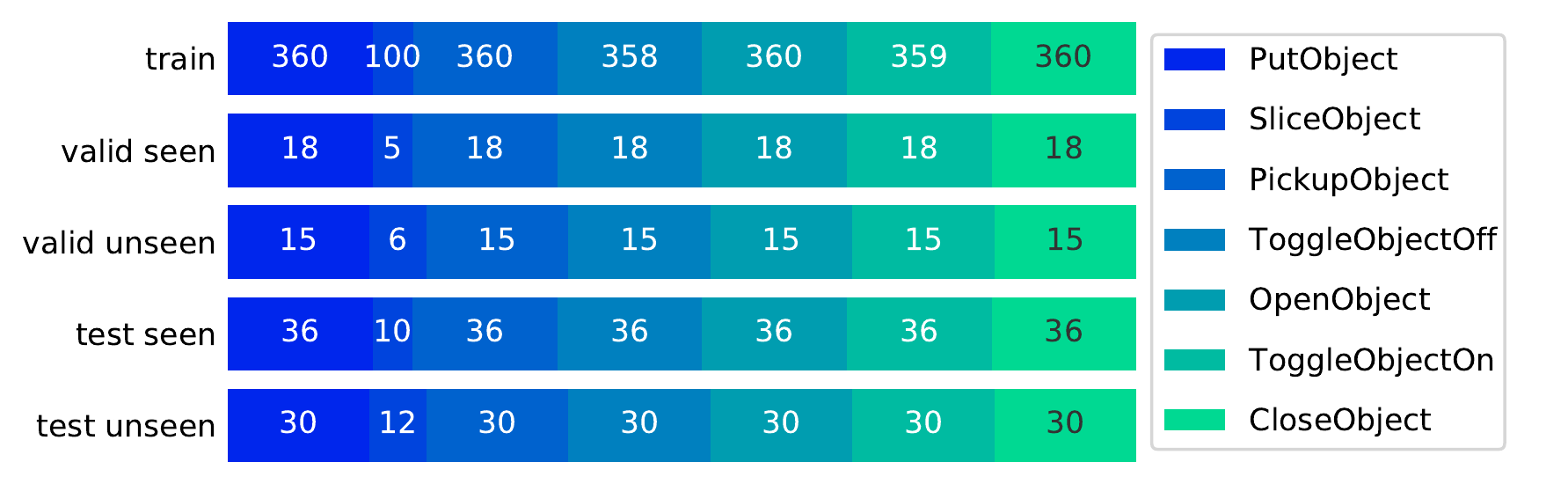}} \\
        \multicolumn{1}{c}{\taskiqa state types} & \\
        \multicolumn{1}{l}{\includegraphics[height=\tasktypeheight]{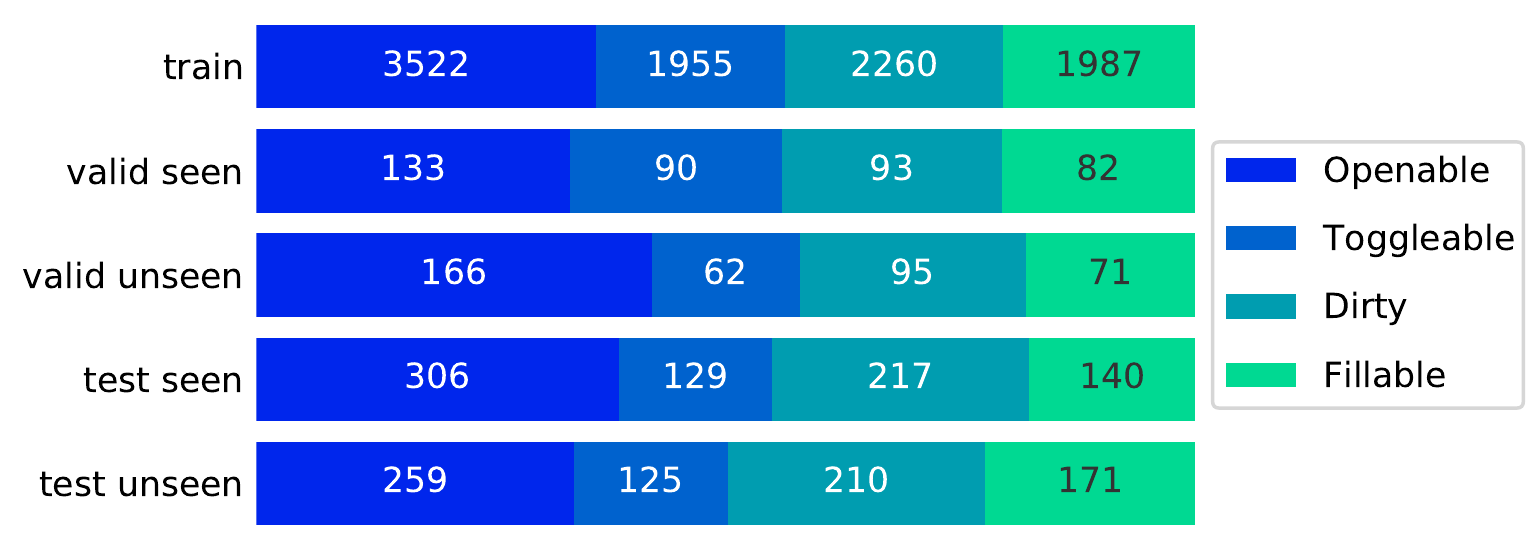}} & \\
    \end{tabular}
    \caption{\textbf{Distribution of task sub-categories} across different splits.}
    \label{fig:task_type}
\end{figure*}

We describe the details of our datasets as mentioned in Section~5 (line 289).
Table~\ref{tab:dataset_template} shows the text templates used to generate the instructions for different tasks. \{obj\}, \{recep\} and \{mrecep\} correspond to target object, receptacle and movable receptacle, respectively. 
Table~\ref{tab:dataset_splits} shows the splits for the four task types, including seen (novel configurations) and unseen (novel scenes) splits for validation and testing. Fig.~\ref{fig:task_type} shows the distribution of sub-categories for each task type (3 for \taskshort and \taskiqa and 7 for \tasklong and \taskexp). In general, the distributions are balanced across sub-categories, but \taskiqa has roughly twice the number of object state questions than for existence or counting. The reason is that there are various types of object states (open/closed, toggled on/off, dirty/clean, or empty/full).

\begin{table*}[h]\footnotesize
    \centering
    \begin{tabular}{lcccccccc}
        \toprule
        & \multicolumn{2}{c}{\taskshort} & \multicolumn{2}{c}{\tasklong} & \multicolumn{2}{c}{\taskiqa} & \multicolumn{2}{c}{\taskexp} \\
        & episodes & scenes & episodes & scenes & episodes & scenes & episodes & scenes \\
        \midrule
        Training     & 2739 & 36 & 8763 & 72 & 19728 & 72 & 2257 & 18 \\
        Valid seen   & 130  & 32 & 350  & 72 & 798   & 72 & 113  & 18 \\
        Valid unseen & 118  & 9  & 350  & 20 & 794   & 20 & 96   & 5  \\
        Test seen    & 349  & 35 & 1050 & 72 & 1592  & 72 & 226  & 18 \\
        Test unseen  & 187  & 10 & 640  & 20 & 1565  & 20 & 192  & 5 \\
        \bottomrule
    \end{tabular}
    \vspace{2pt}
    \caption{{Dataset splits.}}
    \label{tab:dataset_splits}
\end{table*}

\begin{figure*}[h]
    \centering
    \begin{tabular}{cc}
        \tasklong target objects & \tasklong target receptacles \\
        \includegraphics[width=0.45\textwidth]{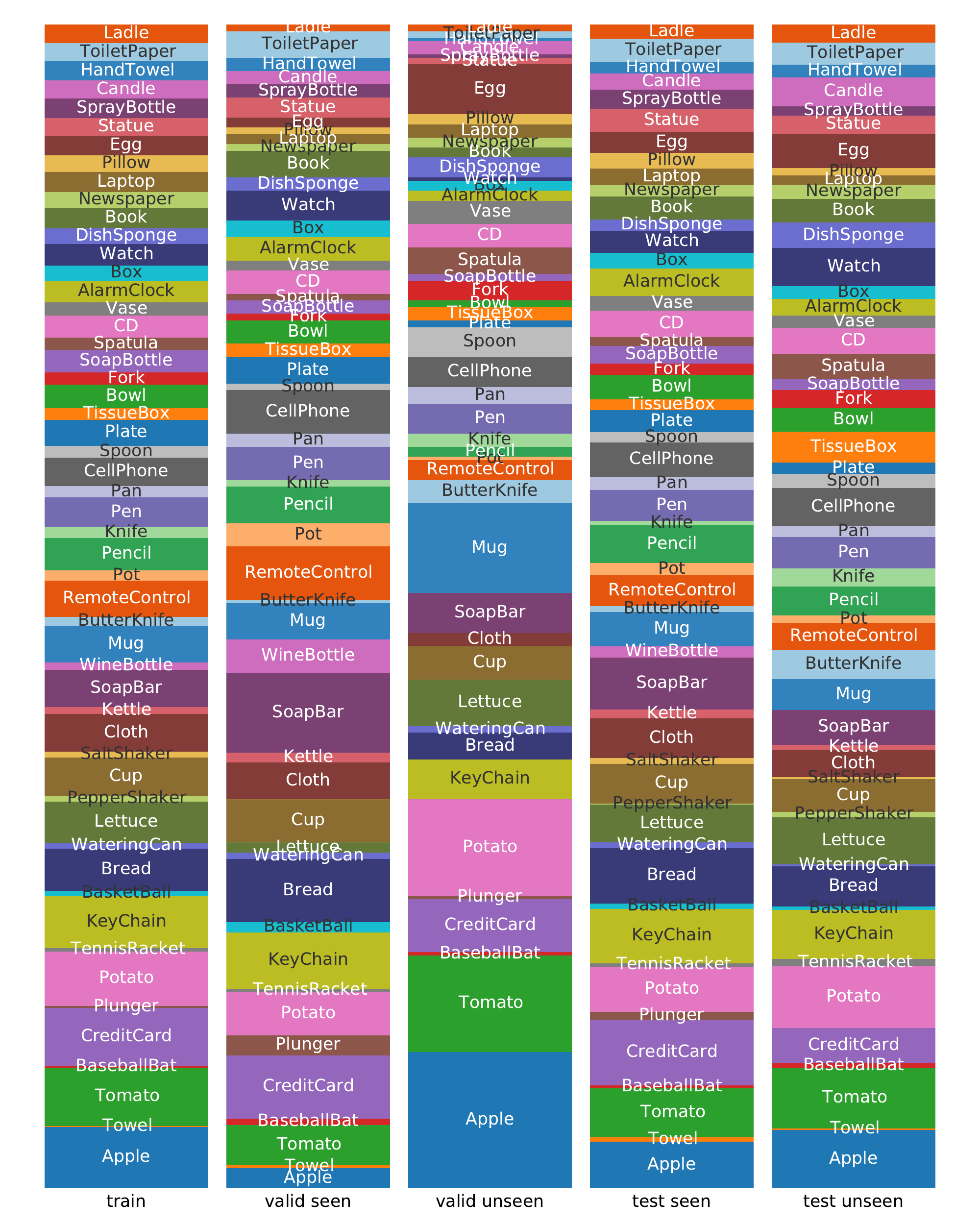} &     \includegraphics[width=0.45\textwidth]{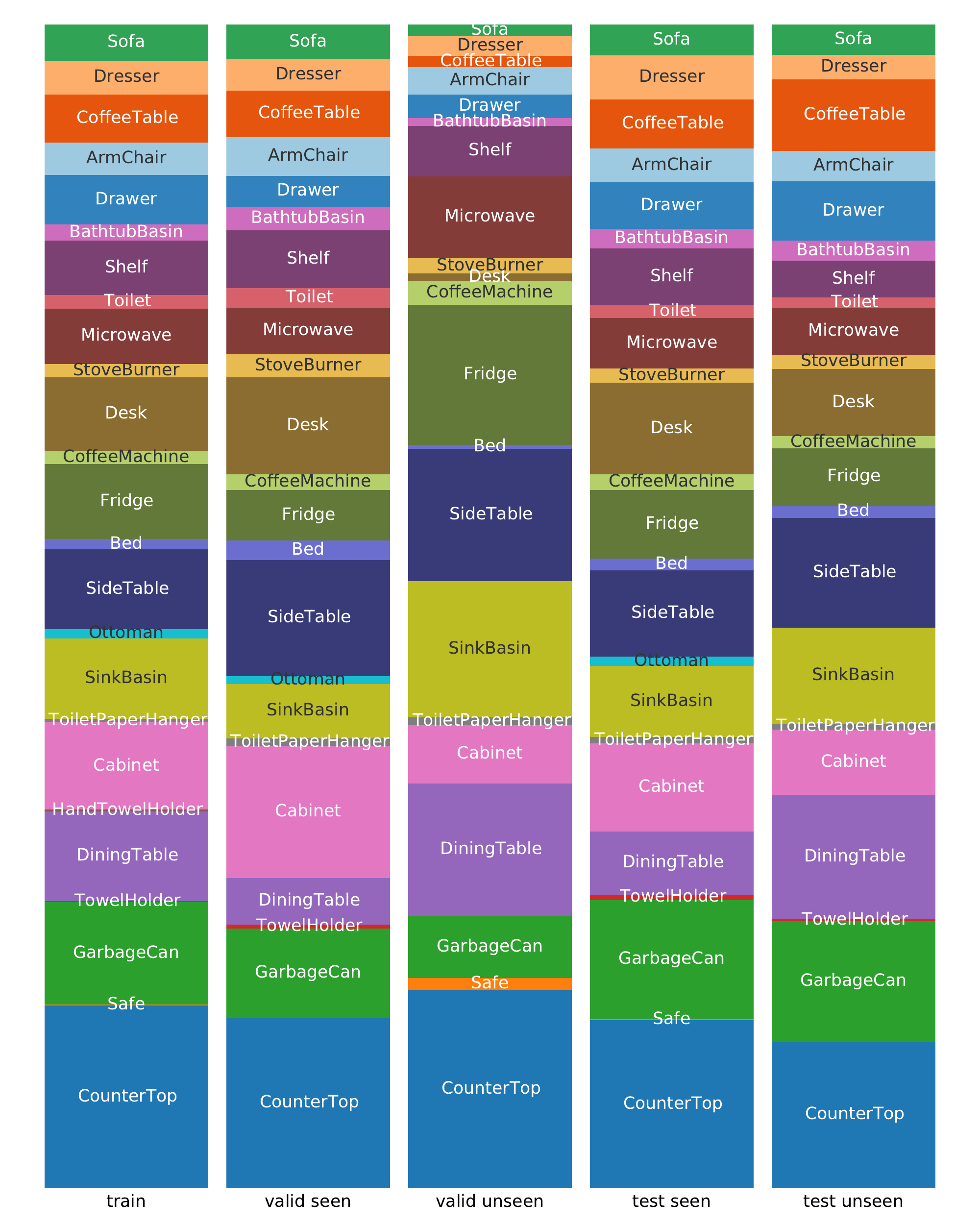} \\
        \taskiqa target objects & \taskiqa target receptacles \\
        \includegraphics[width=0.45\textwidth]{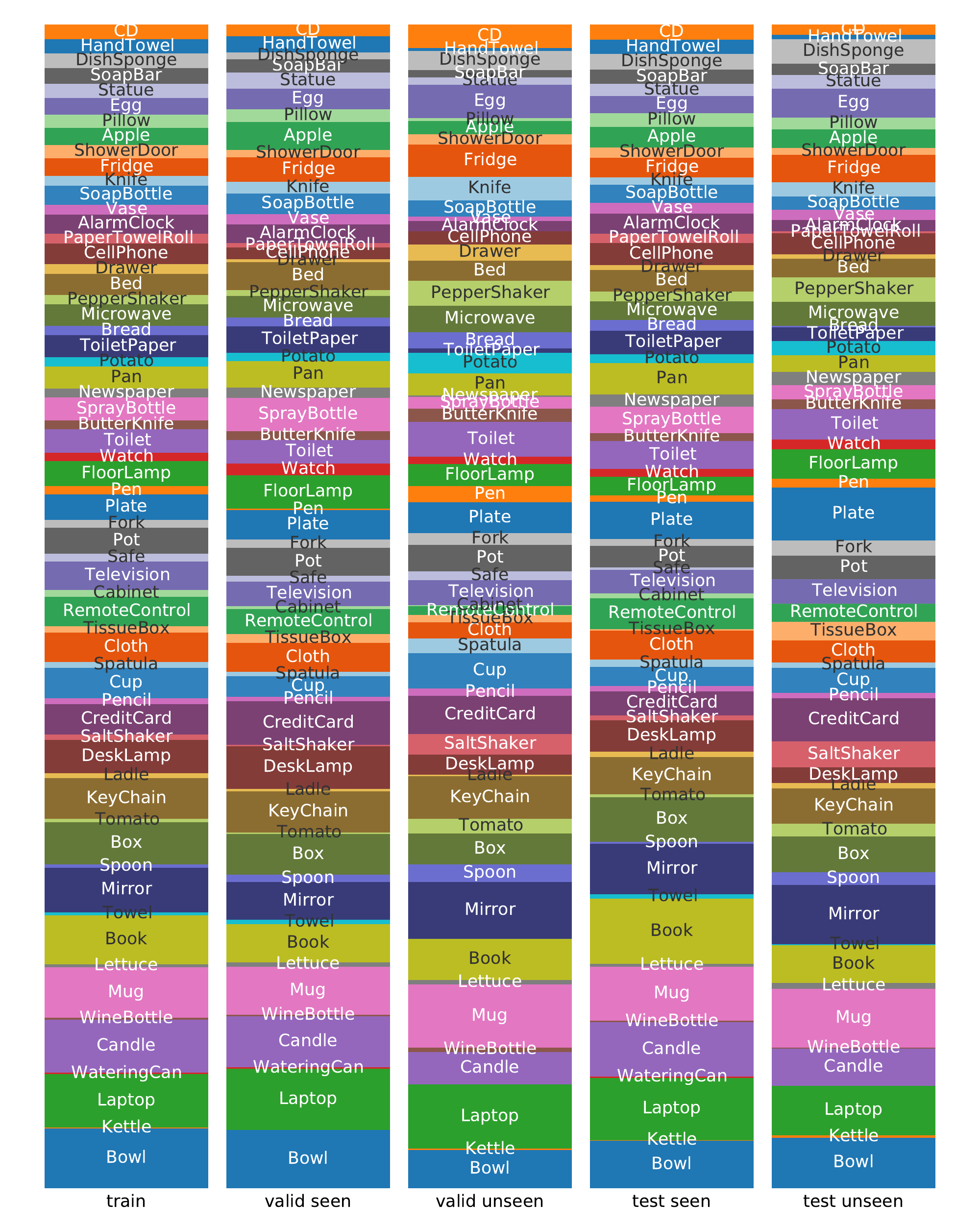} &     \includegraphics[width=0.45\textwidth]{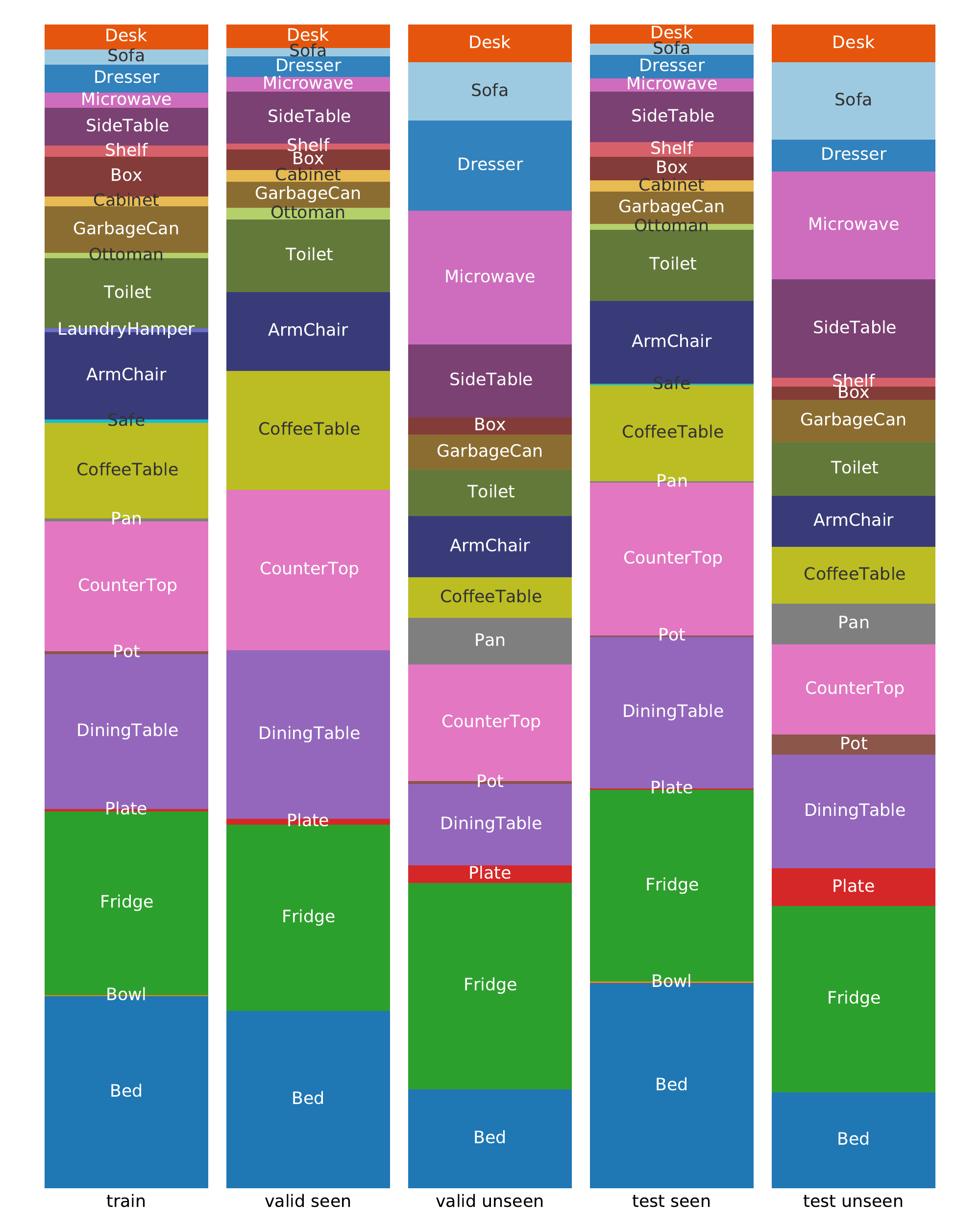} \\
    \end{tabular}
    \caption{\textbf{Target object and receptacle distributions} for \tasklong and \taskiqa splits.}
    \label{fig:distribution_object_receptacle}
\end{figure*}

In order to illustrate the richness in terms of target object types and receptacles in our datasets, Fig.~\ref{fig:distribution_object_receptacle} shows the respective distributions across all five splits in \tasklong and \taskiqa. The other two tasks share a similar distribution. 

\section{More Qualitative Examples}
\begin{figure*}[h]
    \centering
    \includegraphics[width=\textwidth]{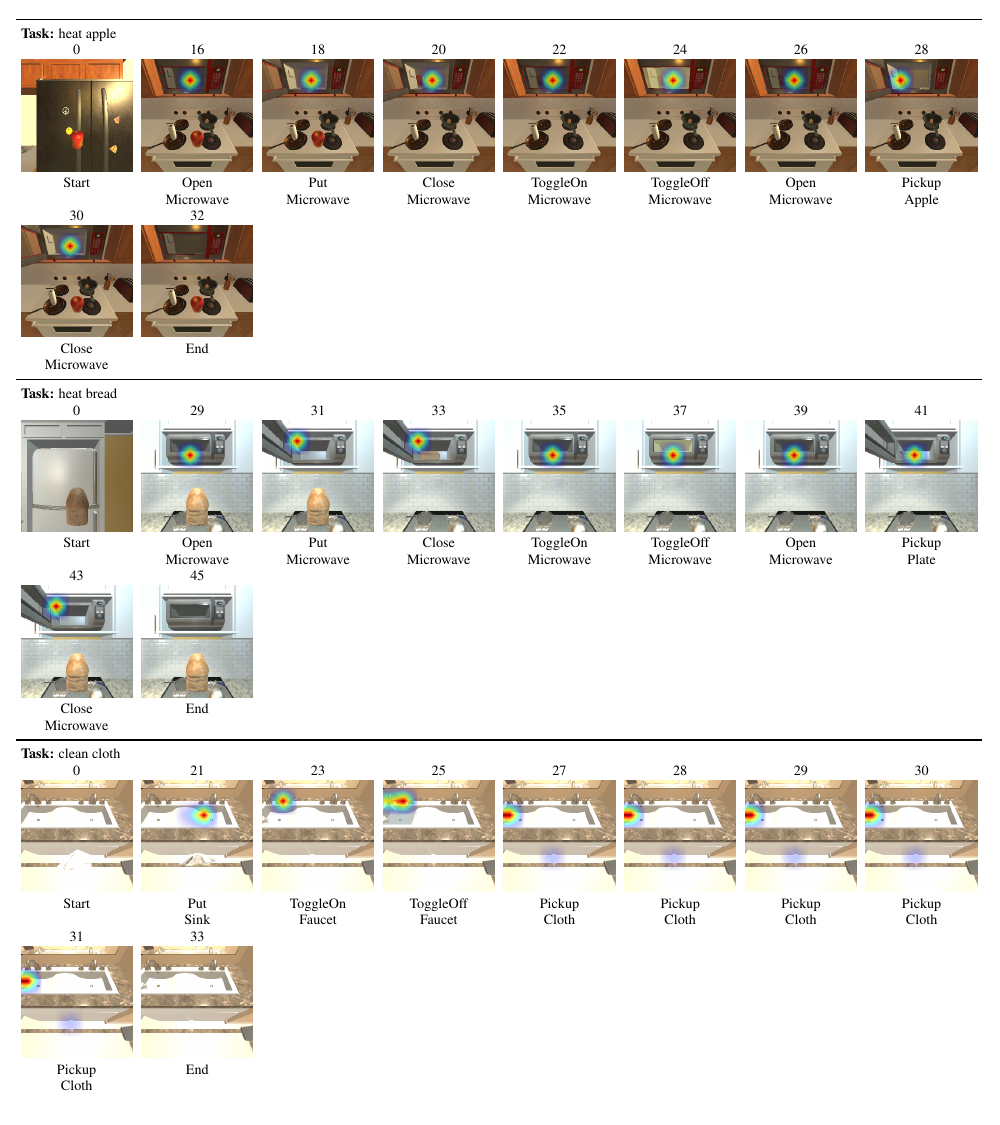}
    \caption{\textbf{Qualitative results of \taskshort.} Valid-Seen trajectories for \taskshort  (part 1), including examples for heating and cleaning in kitchen and bathroom environments. For each example, we show the task instruction, the time step for each frame, overlaid interaction point heat maps for the interaction actions, invoked skill, and the target object for the current skill. Frames not shown correspond to navigation steps.}
    \label{fig:supplement_traj_short1}
\end{figure*}

\begin{figure*}[h]
    \centering
    \includegraphics[width=\textwidth]{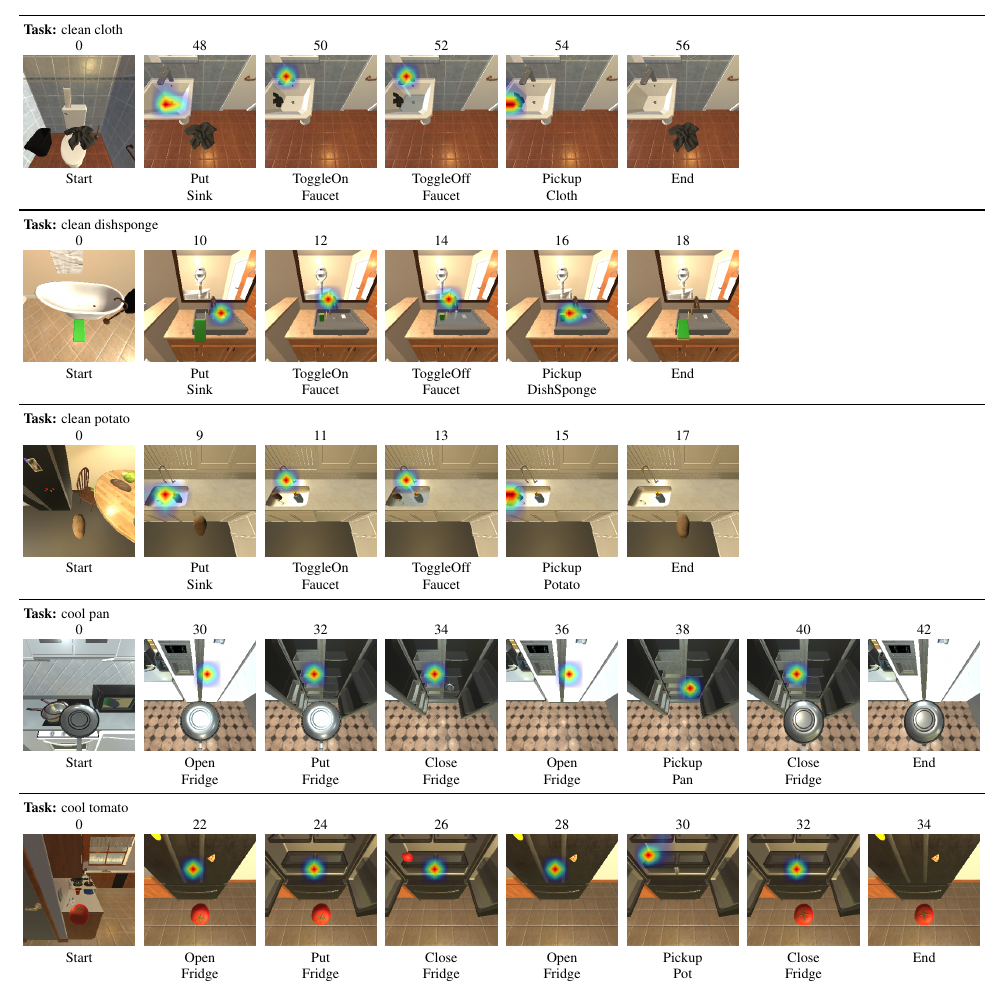}
    \caption{\textbf{Qualitative results of \taskshort.} Valid-Seen trajectories for \taskshort  (part 2), including examples for heating and cleaning in kitchen and bathroom environments. For each example, we show the task instruction, the time step for each frame, overlaid interaction point heat maps for the interaction actions, invoked skill, and the target object for the current skill. Frames not shown correspond to navigation steps.}
    \label{fig:supplement_traj_short2}
\end{figure*}

\begin{figure*}[h]
    \centering
    \includegraphics[width=\textwidth]{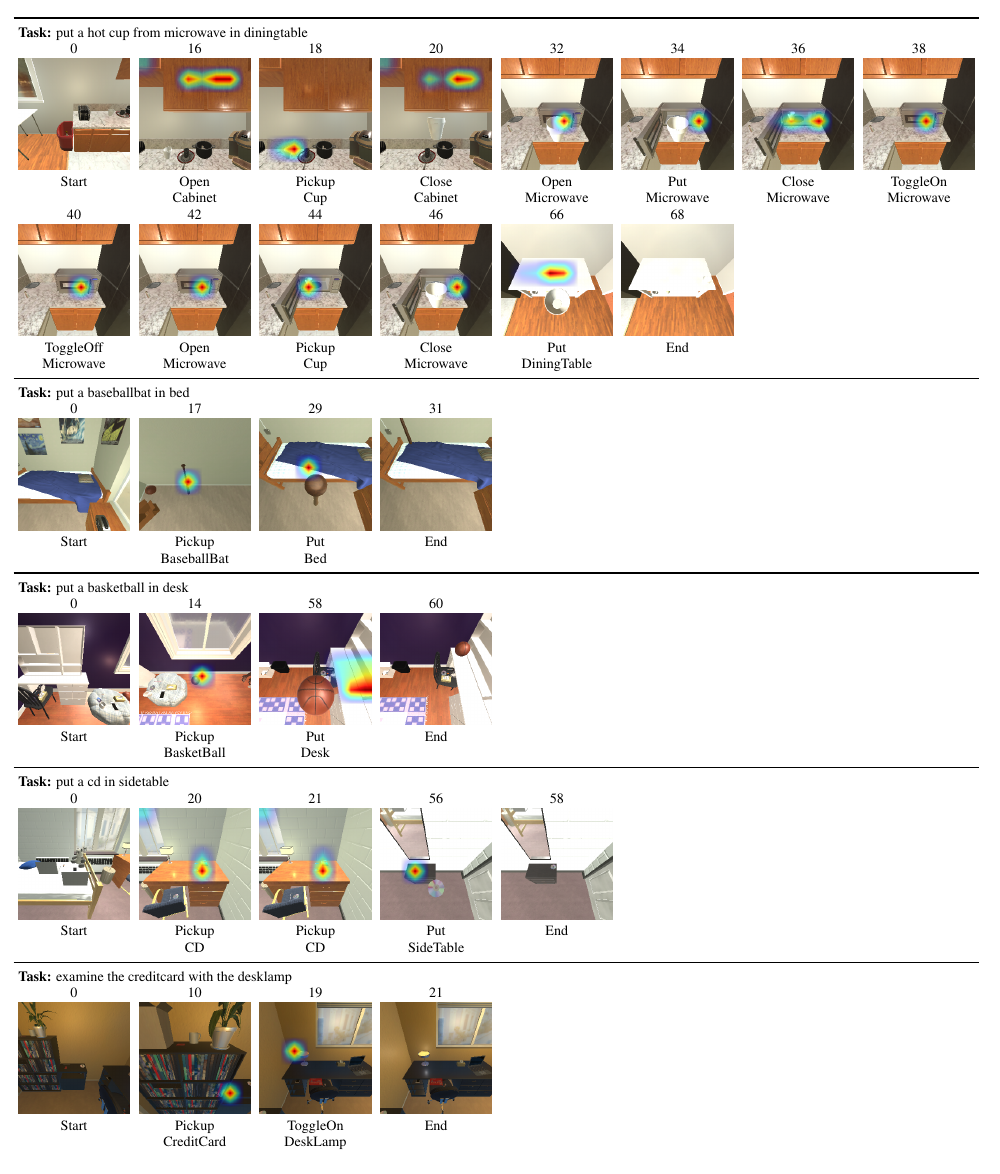}
    \caption{\textbf{Qualitative results of \tasklong.} Valid-Seen trajectories for \tasklong, including examples for picking up, heating and placing on receptacle; as well as shorter tasks involving picking up and placing in receptacle or picking up and examining under (toggled on) light. The environments include kitchen, bedroom, and living room. For each task, we show the task instruction, the time step for each frame, overlaid interaction point heat maps for interactive actions, invoked skill, and the target object for the current skill. Frames not shown correspond to navigation steps.}
    \label{fig:supplement_traj_long}
\end{figure*}

\begin{figure*}[h]
    \centering
    \includegraphics[width=0.85\textwidth]{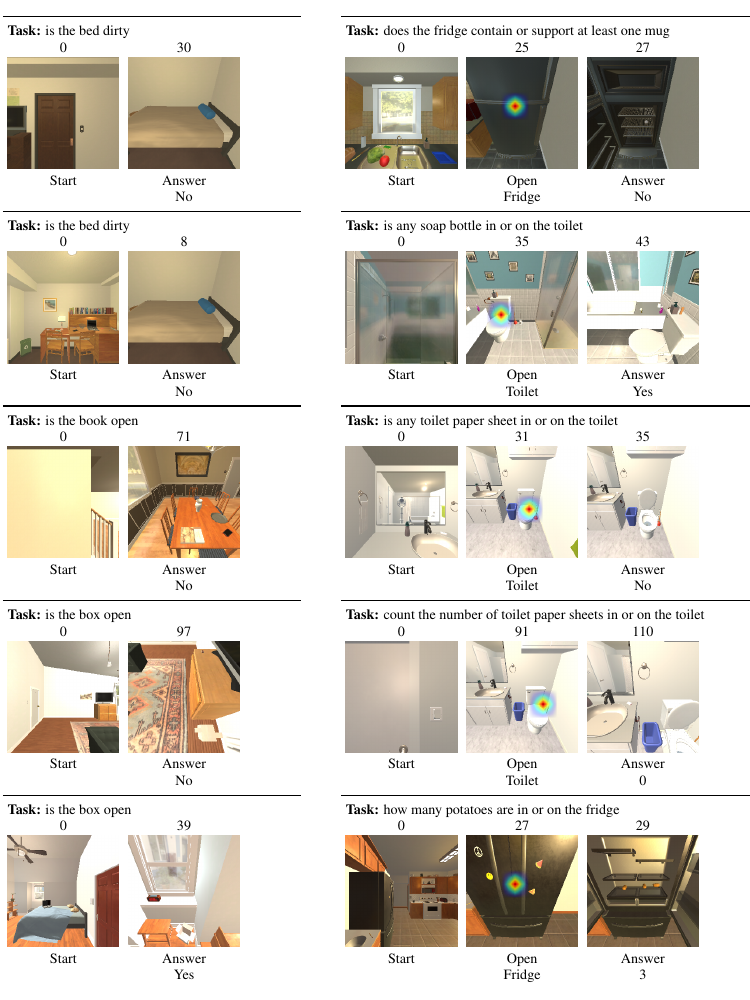}
    \caption{\textbf{Qualitative results of \taskiqa.} Valid-Seen trajectories for \taskiqa, including examples for object state questions, existence and counting in bedrooms, living rooms, kitchens and bathrooms. For each example, we show the question, the time step for each frame, overlaid interaction point heat maps for interaction actions, invoked skill, and the target object for the current skill or the final answer. The two ``is the bed dirty'' examples show two different episode setups for the same question and target object states (note the different initialization). Frames not shown correspond to navigation steps.}
    \label{fig:supplement_traj_iqa}
\end{figure*}

\begin{figure*}[h]
    \centering
    \includegraphics[width=\textwidth]{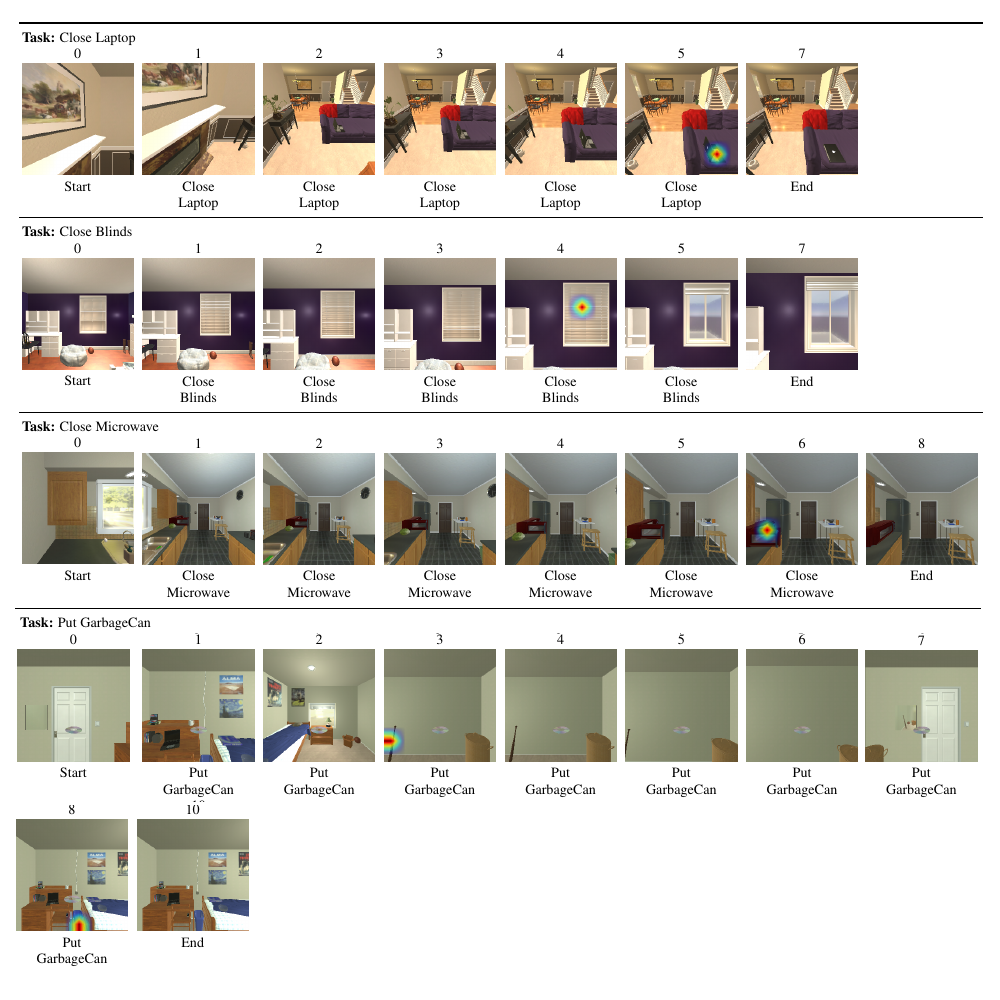}
    \caption{\textbf{Qualitative results of \taskexp.} Valid-Seen trajectories for \taskexp, including examples for closing objects of several scales and placement of an object in a relatively small receptacle in living room, kitchen and bedroom environments. For each example, we show the task instruction, the time step for each frame, overlaid interaction point heat maps for interaction actions, the unique invoked skill, and the corresponding target object.}
    \label{fig:supplement_traj_exp_1}
\end{figure*}

Figs.~\ref{fig:supplement_traj_short1}, \ref{fig:supplement_traj_short2}, \ref{fig:supplement_traj_long}, \ref{fig:supplement_traj_iqa}, and \ref{fig:supplement_traj_exp_1} show several validation \emph{seen} trajectories for all four task types. We only include interaction actions (besides start and end observations) for all tasks.

\begin{figure*}[h]
    \centering
    \includegraphics[width=\textwidth]{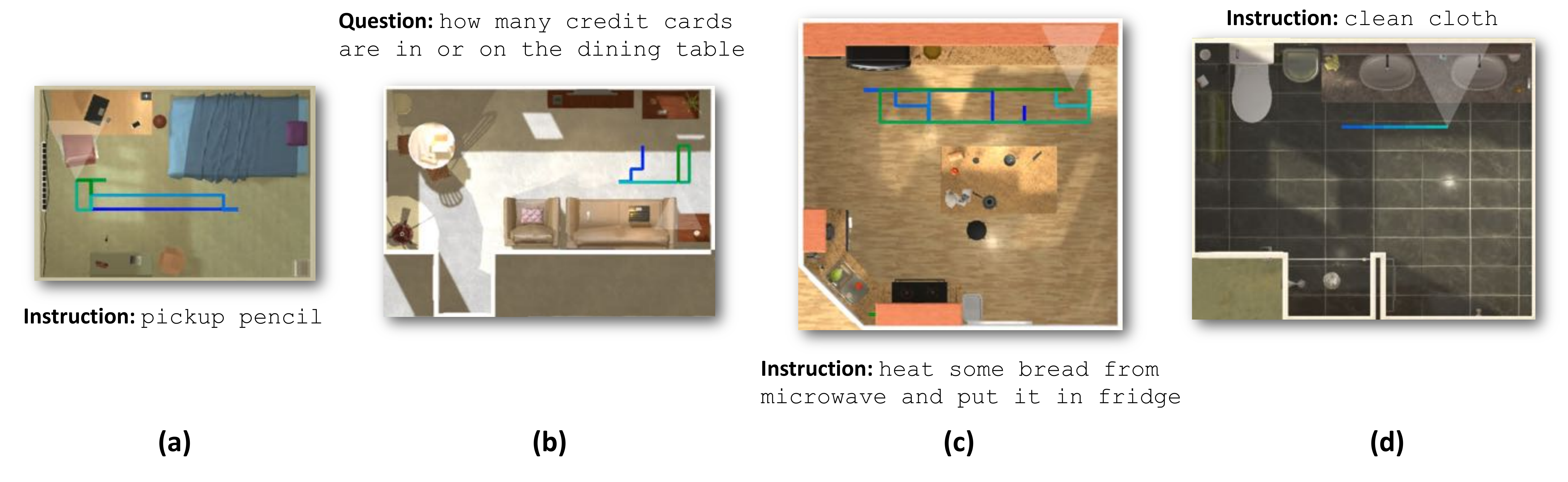}
    \caption{\textbf{Failed episodes in validation-seen.} (a) Picking up small objects like the \emph{pencil} in this \taskexp episode can lead to multiple failed interaction attempts. (b) The model produces its answer in \taskiqa by looking at a \emph{side table} instead of a \emph{dining table}. (c) In this \tasklong episode, the agent keeps exploring the upper part of the kitchen plan and never manages to reach the microwave. (d) Similar to (a), we observe a failed interaction with a small object like the \emph{faucet} in a \taskshort episode.}
    \label{fig:failures}
\end{figure*}

\paragraph{Failures.} Some of the included trajectories show failures that naturally and often occur across all task types and whose accumulation leads to the achieved success rates. Fig.~\ref{fig:supplement_traj_short1} (``heat bread'' and ``clean cloth'') show a wrong target object for the \emph{pickup} skill (\emph{plate} instead of \emph{bread}) issued by the high-level policy, which is recovered by the sub-policy, and failure to pickup an object with a small footprint in the given observation (\emph{cloth}) by the sub-policy. Fig.~\ref{fig:supplement_traj_long} (``put a hot cup from microwave in diningtable'') shows two failed interactions with a \emph{Cabinet}. Fig.~\ref{fig:supplement_traj_iqa} (``is any soap bottle in or on the toilet?'') again shows a failed interaction, which is eventually rendered irrelevant since the answer in this episode could be provided by observing the object lying on the surface of the \emph{toilet}. Fig.~\ref{fig:failures} shows trajectories for failed episodes, where interaction with small or partially occluded objects is a common failure mode.

\section{Dataset Terms of Service}
We use AI2-THOR to create our dataset which is under Apache License 2.0.

\clearpage

\end{document}